\documentclass{article}

    \PassOptionsToPackage{numbers, compress, numbers}{natbib}

    \usepackage[final]{neurips_2020}

\usepackage[utf8]{inputenc} 
\usepackage[T1]{fontenc}    
\usepackage{hyperref}       
\usepackage{url}            
\usepackage{booktabs}       
\usepackage{amsfonts}       
\usepackage{nicefrac}       
\usepackage{microtype}      
\usepackage{xcolor}

\usepackage{amsmath}
\usepackage{algorithm}
\usepackage{algpseudocode}
\usepackage{wrapfig}
\usepackage{lipsum}

\usepackage{graphicx}
\usepackage[normalem]{ulem}
\usepackage{caption}
\usepackage{subcaption}
\usepackage{appendix}
\usepackage{relsize}

\usepackage{calc}
\newcommand{\explain}[2]{\underbrace{#1}_{\parbox{\widthof{\ensuremath{#1}}}{\footnotesize\raggedright #2}}}

\newcommand{\mujocobaselinefigsize}{0.143}

\newcommand{\s}{\sigma}

\newcommand{\w}{\omega}

\newcommand{\N}{\mathcal{N}}

\DeclareMathOperator*{\argmin}{argmin}
\DeclareMathOperator*{\argmax}{argmax}

\usepackage[export]{adjustbox}

\title{Multi-task Batch Reinforcement Learning with Metric Learning}

\newcommand*\samethanks[1][\value{footnote}]{\footnotemark[#1]}

\author{
    Jiachen Li$^{1}$\thanks{Equal Contribution}
    \quad Quan Vuong$^{1}$\samethanks\
    \quad Shuang Liu$^{1}$
    \quad Minghua Liu$^{1}$ \\
    \quad \textbf{Kamil Ciosek}$^{2}$ 
    \quad \textbf{Keith Ross}$^{3, 4}$
    \quad \textbf{Henrik Christensen}$^{1}$ 
    \quad \textbf{Hao Su}$^{1}$ \\
    {${ }^{1}$UC San Diego \qquad
    ${ }^{2}$ Microsoft Research Cambridge, UK
    }\\
    {${ }^{3}$ New York University Shanghai \qquad ${ }^{4}$ New York University 
    }\\
    
  {\texttt{\{jil021, qvuong\}@ucsd.edu}} \\
}

\begin{document}

\maketitle

\begin{abstract}
  We tackle the Multi-task Batch Reinforcement Learning problem. Given multiple datasets collected from different tasks, we train a multi-task policy to perform well in unseen tasks sampled from the same distribution. The task identities of the unseen tasks are not provided. To perform well, the policy must infer the task identity from collected transitions by modelling its dependency on states, actions and rewards. Because the different datasets may have state-action distributions with large divergence, the task inference module can learn to ignore the rewards and spuriously correlate \textit{only} state-action pairs to the task identity, leading to poor test time performance. To robustify task inference, we propose a novel application of the triplet loss. To mine hard negative examples, we relabel the transitions from the training tasks by approximating their reward functions. When we allow further training on the unseen tasks, using the trained policy as an initialization leads to significantly faster convergence compared to randomly initialized policies (up to $80\%$ improvement and across 5 different Mujoco task distributions). We name our method \textbf{MBML} (\textbf{M}ulti-task \textbf{B}atch RL with \textbf{M}etric \textbf{L}earning) \footnote{Website: \url{https://sites.google.com/eng.ucsd.edu/multi-task-batch-reinforcement/home}}.

\end{abstract}

\section{Introduction}\label{intro}

Combining neural networks (NN) with reinforcement learning (RL) has led to many recent advances \cite{SOP, PPO, SAC, TRPO, SPU}.
Since training NNs requires diverse datasets and collecting real world data is expensive, most RL successes are limited to scenarios where the data can be cheaply generated in a simulation.
On the other hand, offline data is essentially free for many applications and RL methods should use it whenever possible. 
This is especially true because practical deployments of RL are bottle-necked by its poor sample efficiency.
This insight has motivated a flurry of recent works in Batch RL \cite{siegel2020keep,agarwal2019optimistic,kumar2019stabilizing,fujimoto2019off,chen2019bail}. 
These works introduce specialized algorithms to stabilize training from offline datasets. 
However, offline datasets are not necessarily diverse. 
In this work, we investigate how the properties of a diverse dataset influence the policy search procedure.
By collecting diverse offline dataset, we hope the networks will generalize without further training to unseen tasks or provide good initialization that speeds up convergence when we perform further on-policy training.

To collect diverse datasets, it occurs to us that we should collect data from different tasks. 
However, datasets collected from different tasks may have state-action distributions with large divergence.
Such dataset bias presents a unique challenge in robust task inference. 
We provide a brief description of the problem setting, the challenge and our contributions below. 
For ease of exposition, we refer to such datasets as having little overlap in their state-action visitation frequencies thereafter.

\begin{figure}[t]
    \centering
    \includegraphics[width=0.5\textwidth]{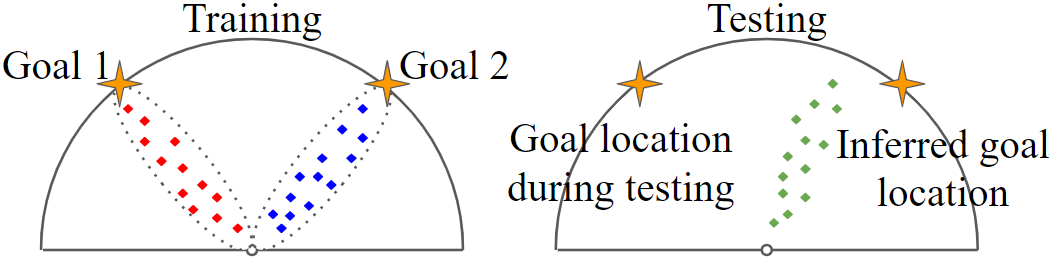}
    \caption{A toy example to illustrate the challenge. The agent must navigate from the origin to a goal location. \textbf{Left:} Goal 1 and Goal 2 denote the two training tasks. The red and blue squares indicate the transitions collected from task 1 and 2 respectively. We can train the task inference module to infer the task identity to be 1 when the context set contains the red transitions and 2 when the context set contains the blue transitions. Since there are no overlap between the red and blue squares, the task inference module learns to correlate the state-action pairs to the task identity. \textbf{Right:} The failure of the task inference module. The policy must infer the task identity from the randomly collected transitions, denoted by the green squares.
    The agent needs to navigate to goal 1 during testing. However, if the green squares have more overlap with the blue squares, the task inference module will predict 2 to be the task identity. The agent therefore navigates to the wrong goal location.}
    \label{fig:cheat}
\end{figure}

We tackle the Multi-task Batch RL problem. 
We train a policy from multiple datasets, each generated by interaction with a different task. 
We measure the performance of the trained policy on unseen tasks sampled from the same task distributions as the training tasks. 
To perform well, the policy must first infer the identity of the unseen tasks from collected transitions and then take the appropriate actions to maximize returns. 
To train the policy to infer the task identity, we can train it to distinguish between the different training tasks when given transitions from the tasks as input. 
These transitions are referred to as the context set \cite{rakelly2019efficient}. 
Ideally, the policy should model the dependency of the task identity on both the rewards and the state-action pairs in the context set. 
To achieve this, we can train a task identification network that maps the collected experiences, including both state-action pairs and rewards, to the task identity or some task embedding. 
This approach, however, tends to fail in practice. 
Since the training context sets do not overlap significantly in state-action visitation frequencies, it is possible that the learning procedure would minimize the loss function for task identification by \textit{only} correlating the state-action pairs and ignoring rewards, which would cause mistakes in identifying testing tasks. 
This is an instance of the well-known phenomena of ML algorithms cheating when given the chance \cite{chu2017cyclegan} and is further illustrated in Fig. \ref{fig:cheat}. 
We limit our explanations to the cases where the tasks differ in reward functions. Extending our approach to task distribution with different transition functions is easily done. We provide experimental results for both cases.

Our contributions are as follows. 
To the best of our knowledge, we are the first to highlight the issue of the task inference module learning the wrong correlation from biased dataset. 
We propose a novel application of the triplet loss to robustify task inference.
To mine hard negative examples, we approximate the reward function of each task and relabel the rewards in the transitions from the other tasks. 
When we train the policy to differentiate between the original and relabelled transitions, we force it to consider the rewards since their state-action pairs are the same. 
Training with the triplet loss generalizes better to unseen tasks compared to alternatives. 
When we allow further training on the unseen tasks, using the policy trained from the offline datasets as initialization significantly increase convergence speed (up to $80\%$ improvement in sample efficiency).

To the best of our knowledge, the most relevant related work is \cite{siegel2020keep}, which is solving a different problem from ours. 
They assume access to the ground truth task identity and reward function of the testing task. 
Our policy does not know the testing task's identity and must infer it through collected trajectories. 
We also do not have access to the reward function of the testing tasks.

\section{Preliminaries and Problem Statement}
\label{sec_prelim}

\textbf{To help the reader follow our explanation, we include a symbol definition table in \autoref{sec_def_var}}.







We model a task as a Markov Decision Process $M = (\mathcal{S}, \mathcal{A}, T, T_0, R, H)$, with state space $\mathcal{S}$, action space $\mathcal{A}$, transition function $T$, initial state distribution $T_0$, reward function $R$, and horizon $H$. At each discrete timestep $t$, the agent is in a state $s_t$, picks an action $a_t$, arrives at $s'_t\sim T(\cdot|s_t, a_t)$, and receives a reward $R(s_t, a_t, s'_t)$. The performance measure of policy $\pi$ is the expected sum of rewards $J_M(\pi) = \mathbb{E}_{\tau_M \sim \pi}[\sum_{t=0}^{H-1} R(s_t, a_t, s'_t)]$, where $\tau_M = (s_0, a_0, r_0, s_1, a_1, r_1, \ldots)$ is a trajectory generated by using $\pi$ to interact with $M$.

\subsection{Batch Reinforcement Learning}\label{BCQ}
A Batch RL algorithm solves the task using an existing batch of $N$ transitions $\mathcal{B} = \{(s_{t}, a_{t}, r_{t}, s'_{t}) | t = 1, \ldots, N\}$. A recent advance in this area is Batch Constrained Q-Learning (BCQ) \cite{fujimoto2019off}. Here, we explain how BCQ selects actions. Given a state $s$, a generator $G$ outputs multiple candidate actions $\{a_m\}_{m}$. A perturbation model $\xi$ takes as input the state-candidate action and generates small correction $\xi(s, a_m)$. The corrected action with the highest estimated $Q$ value is selected as $\pi\left(s\right)$:
\begin{equation}
    \pi\left(s\right)=\underset{a_{m}+\xi\left(s, a_{m}\right)}{\arg \max } Q\left(s, a_{m}+\xi\left(s, a_{m}\right)\right), \quad \quad\left\{a_{m}=G\left(s, \nu_{m}\right)\right\}_{m}, \quad \quad \nu_{m} \sim \mathcal{N}(0,1).
\end{equation}
To help the reader follow our discussion, we illustrate graphically how BCQ selects action in \autoref{sec_action_bcq}. 
In our paper, we use BCQ as a routine. The take-away is that BCQ takes as input a batch of transitions $\mathcal{B} = \{(s_{t}, a_{t}, r_{t}, s'_{t}) | t = 1, \ldots, N\}$ and outputs three learned functions $Q, G, \xi$.
\subsection{Multi-task Batch Reinforcement Learning}
\label{bg_mtbrl}

Given $K$ batches, each containing $N$ transition tuples from one task,
$\mathcal{B}_{i}=\{(s_{i, t}, a_{i, t}, r_{i, t}, s_{i, t}^{\prime}) | i = 1, \ldots, K, t=1, \ldots, N \}$,
we define the Multi-task Batch RL problem as: 
\begin{equation}\label{eq_obj}
\underset{\theta}{\arg \max } \,\, J(\theta)=\mathbb{E}_{M_{i} \sim p\left(M\right)}\left[ J_{M_i}(\pi_\theta)\right],
\end{equation}
where an algorithm only has access to the $K$ batches and $J_{M_i}(\pi)$ is the performance of the policy $\pi$ in task $i$, i.e. $\mathbb{E}_{\tau_{M_i} \sim \pi}[\sum_{t=0}^{H-1} R(s_{i, t}, a_{i, t}, s'_{i, t})]$. $p(M)$ defines a task distribution. The subscript $i$ indexes the different tasks. The tasks 
have the same state and action space and
only differ in the transition and reward functions \cite{zintgraf2020varibad}. A distribution over the transition and/or the reward functions therefore defines the task distribution. We measure performance by computing average returns over unseen tasks sampled from the same task distribution. The policy is not given identity of the unseen tasks before evaluation and must infer it from collected transitions.

In multi-task RL, we can use a task inference module $q_\phi$ to infer the task identity from a context set. The context set for a task $i$ consists of transitions from task $i$ and is denoted ${\bf c}_i$.
The task inference module $q_\phi$ takes ${\bf c}_i$ as input and outputs a posterior over the task identity. We sample a task identity ${\bf z}_i$ from the posterior and inputs it to the policy in addition to the state, i.e. $\pi(s, {\bf z}_i)$. 
We model $q_\phi$ with the probabilistic and permutation-invariant architecture from \cite{rakelly2019efficient}. $q_\phi$ outputs the parameters of a diagonal Gaussian. For conciseness, we sometimes use the term policy to also refer to the task inference module. It should be clear from the context whether we are referring to $q_\phi$ or $\pi$.

We evaluate a policy on unseen tasks in two different scenarios: (1) Allowing the policy to collect a small number of interactions to infer $z$, we evaluate returns without further training, (2) Training the policy in the unseen task and collecting as much data as needed, we evaluate the amount of transitions the policy needs to collect to converge to the optimal performance.
 
We assume that each batch $\mathcal{B}_{i}$ contains data generated by a policy while learning to solve task $M_{i}$. Thus, if solving each task involve visiting different subspace of the state space, the different batches do not have significant overlap in their state-action visitation frequencies. This is illustrated in Fig. \ref{fig:cheat}.

\section{Proposed algorithm}
\label{sec_algo}

\subsection{Learning multi-task policy from offline data with distillation}\label{sec_algo_distillation}

In Multi-task RL, \cite{rusu2015policy, teh2017distral, ghosh2017divide,czarnecki2019distilling, ActorMimicParisotto2015} demonstrate the success of distilling multiple single-task policies into a multi-task policy. Inspired by these works, we propose a distillation procedure to obtain a multi-task policy in the Multi-task Batch RL setting. In Sec. \ref{sec_algo_triplet}, we argue such distillation procedure alone is insufficient due to the constraints the batch setting imposes on the policy search procedure.

The distillation procedure has two phases. 
In the first phase, we use BCQ to learn a different policy for each task, i.e. we learn $K$ different and independent policies. While we can use any Batch RL algorithm in the first phase, we use BCQ due to its simplicity.
As described in Sec. \ref{BCQ}, 
for each training batch, BCQ learns three functions: a state-action value function $Q$, a candidate action generator $G$ and a perturbation generator $\xi$. The output of the first phase thus consists of three sets of networks $\{Q_i\}^K_{i=1}$, $\{G_i\}^K_{i=1}$, and $\{\xi_i\}^K_{i=1}$, where $i$ indexes over the training tasks.

In the second phase, we distill each set into a network by incorporating a task inference module. The distilled function should recover different task-specific function depending on the inferred task identity. To distill the value functions $\{Q_i\}^K_{i=1}$ into a function $Q_D$, for each task $i$, we sample a context ${\bf c}_{i}$ and a pair $(s, a)$ from the batch $\mathcal{B}_i$. The task inference module $q_\phi$ takes ${\bf c}_{i}$ as input and infers a task identity ${\bf z}_i$. Given ${\bf z}_i$ as input, $Q_D$ should assign similar value to $(s, a)$ as the value function for the $i^{th}$ task $Q_i(s, a)$. The loss function with a $\beta$-weighted KL term \cite{rakelly2019efficient} is:
\begin{equation}\label{eq_loss_qi}
    \mathcal{L}_Q = \frac{1}{K} \sum_{i=1}^{K} \mathlarger{\underset{\substack{(s, a),  {\bf c}_{i} \sim  \mathcal{B}_i }}{\mathbb{E}} }\left[( Q_i(s, a) - Q_D(s, a, {\bf z}_i) )^2 + \beta\text{KL}(q_\phi({\bf c}_i) ||  \mathcal{N}(0,1))\right], \, {\bf z}_i \sim q_\phi({\bf c}_{i})
\end{equation}
We also use Eq. \ref{eq_loss_qi} to train $q_\phi$ using the reparam trick \cite{kingma2013auto}. Similarly, we distill the candidate action generators $\{G_i\}^K_{i=1}$ into $G_D$. $G_D$ takes as input state $s$, random noise $\nu$ and task identity ${\bf z}_i$. Depending on ${\bf z}_i$'s value, we train $G_D$ to regress towards the different candidate action generator:
\begin{equation}\label{eq_loss_gi}
    \mathcal{L}_G = \frac{1}{K} \sum_{i=1}^{K} \mathlarger{\underset{\substack{s, {\bf c}_{i} \sim \mathcal{B}_i \\ \nu \sim \mathcal{N}(0,1) }}{\mathbb{E}} } \left[ || G_i(s, \nu) - G_D(s, \nu, \bar{\bf z}_i) || ^ 2 \right], \quad {\bf z}_i \sim q_\phi({\bf c}_{i}).
\end{equation}
The bar on top of $\bar{\bf z}_i$ in Eq. \ref{eq_loss_gi} indicates the stop gradient operation. We thus do not use the gradient of Eq. \ref{eq_loss_gi} to train the task inference module \cite{rakelly2019efficient}. Lastly, we distill the perturbation generators $\{\xi_i\}^K_{i=1}$ into a single network $\xi_D$ (Eq. \ref{eq_loss_xi_i}). 
$\xi_D$ takes as input a state $s$, a candidate action $a$, and an inferred task identity ${\bf z}_i$. We train $\xi_D$ to regress towards the output of $\xi_i$ given the same state $s$ and candidate action $a$ as input. 
We obtain the candidate action $a$ by passing $s$ through the candidate action generator $G_i$.
\begin{equation}\label{eq_loss_xi_i}
  \mathcal{L}_\xi = \frac{1}{K} \sum_{i=1}^{K} \mathlarger{\underset{\substack{s, {\bf c}_{i} \sim \mathcal{B}_i \\ \nu \sim \mathcal{N}(0,1)} }{\mathbb{E}}}\left[ || \xi_i(s, a) - \xi_D(s, a, \bar{\bf z}_i) || ^ 2 \right], \quad {\bf z}_i \sim q_\phi({\bf c}_{i}), \quad a = G_i(s, \nu)
\end{equation}
Note that the gradient of $\mathcal{L}_\xi$ also updates $G_i$. The final distillation loss is given in Eq. \ref{eq_distill_loss}. We parameterize $q_\phi, Q_D, G_D, \xi_D$ with feedforward NN as detailed in Appendix \ref{sec_hp_distillation}.
\begin{equation}\label{eq_distill_loss}
   \mathcal{L}_{distill} = \mathcal{L}_Q + \mathcal{L}_G + \mathcal{L}_\xi.
\end{equation}

\subsection{Robust task inference with triplet loss design}\label{sec_algo_triplet}
\begin{wrapfigure}{R}{0.269\textwidth}
    \centering
    \vspace{-0.5em}
    \begin{subfigure}{0.145\paperwidth}
    \includegraphics[width=\linewidth]{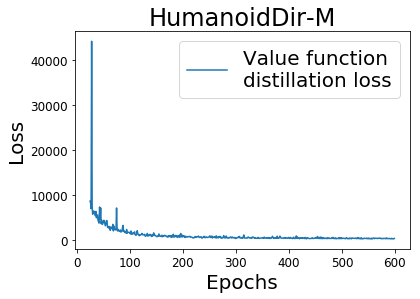}
  \end{subfigure}
  
  \begin{subfigure}{0.145\paperwidth}
    \includegraphics[width=\linewidth]{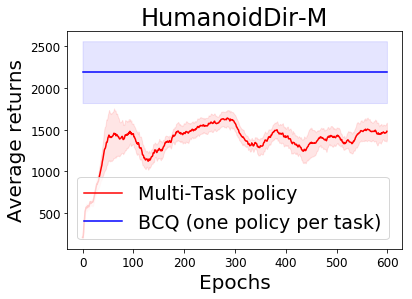}
  \end{subfigure}
  \caption{{\bf Top:} Value function
   distillation
  loss (Eq. \ref{eq_loss_qi}) during training. {\bf Bottom:} The performance of the multi-task policy trained with Eq. \ref{eq_distill_loss} versus BCQ.}\label{fig:fail-distill}
\end{wrapfigure}

Given the high performance of distillation in Multi-task RL \cite{rusu2015policy, teh2017distral, ghosh2017divide,czarnecki2019distilling, ActorMimicParisotto2015}, it surprisingly performs poorly in Multi-task Batch RL, even on the training tasks. This is even more surprising because we can minimize the distillation losses (Fig. \ref{fig:fail-distill} top) and the single-task BCQ policies have high performance (Fig. \ref{fig:fail-distill} bottom). If the single-task policies perform well and we can distill them into a multi-task policy, why does the multi-task policy have poor performance? We argue the task inference module has learnt to model the posterior over task identity as conditionally dependent on only the state-action pairs in the context set
, i.e. $P(Z|S, A)$,
where $S, A$ are random variables denoting states and actions, rather than the correct dependency $P(Z|S, A, R)$ where $R$ denotes the rewards.

The behavior of the trained multi-task policy supports this argument. In this experiment, each task corresponds to a running direction. To maximize returns, the policy should run with maximal velocity in the target direction. We found that the multi-task policy often runs in the wrong target direction, indicating incorrect task inference.
At the beginning of evaluation, the task identity is not provided.
The policy takes random actions, after which it uses the collected transitions to infer the task identity. 
Having learnt the wrong conditional dependency, the task inference module assigns high probability mass in the posterior to region in the task embedding space whose training batches overlap with the collected transitions (Fig. \ref{fig:cheat}).

The fundamental reason behind the wrong dependency is the non-overlapping nature of the training batches.
Minimizing the distillation loss does not require the policy to learn the correct but more complex dependency. 
The multi-task policy should imitate different single-task policy depending on which batch the context set was sampled from.
If the batches do not overlap in state-action visitation frequencies, the multi-task policy can simply correlate the state-action pairs in the context with which single-task policy it should imitate. In short, if minimizing the training objective on the given datasets does not require the policy to model the dependency of the task identity on the rewards in the context set, there is no guarantee the policy will model this dependency. This is not surprising given literature on the non-identifiability of causality from observations \cite{pearl_2009, Peters2017}. They also emphasize the benefit of using distribution change as training signal to learn the correct causal relationship \cite{Bengio2020A}.

Inspired by this literature, we introduce a distribution change into our dataset by approximating the reward function of each task $i$ with a learned function $\hat{R}_i$ (training illustrated in \autoref{sec_ensemble}). Given a context set ${\bf c}_j$ from task $j$, we relabel the reward of each transition in ${\bf c}_j$ using $\hat{R}_i$. Let $t$ index the transitions and ${\bf c}_{j\rightarrow{i}}$ denote the set of the relabelled transitions, we illustrate this process below :
\begin{equation}\label{eq_relabel}
\quad {\bf c}_{j}=\left\{ \left(s_{j, t}, a_{j, t}, r_{j, t}, s_{j, t}^{\prime}\right) \right\}_t   \xrightarrow{\text{Relabelling}} {\bf c}_{j\rightarrow{i}} = \left\{ \left(s_{j, t}, a_{j, t}, \hat{R}_i(s_{j, t}, a_{j, t}), s_{j, t}^{\prime} \right) \right\}_t
\end{equation}
Given the relabelled transitions, we leverage the triplet loss from the metric learning community \cite{hermans2017defense} to enforce robust task inference, which is the most important design choice in MBML. Let $K$ be the number of training tasks, ${\bf c}_i$ be a context set for task $i$, ${\bf c}_j$ be 
a context set for task $j$ ($j \neq i$)
, and ${\bf c}_{j\rightarrow{i}}$ be the relabelled set as described above, the triplet loss for task $i$ is:
\begin{equation}\label{eq_triplet_task_i}
    \mathcal{L}^i_{triplet} = \frac{1}{K-1} \sum_{j=1, j \neq i }^{K} \bigg[\explain{d \big(q_{\phi}\left({\bf c}_{j\rightarrow{i}}\big), q_{\phi}\left({\bf c}_{i}\right)\right) \quad}{Ensure ${\bf c}_{j\rightarrow{i}}$ and ${\bf c}_{i}$ infer \textit{similar} task identities \quad}
    - \explain{ \quad d\big(q_{\phi}\left({\bf c}_{j\rightarrow{i}}\right), q_{\phi}\left({\bf c}_{j}\right)\big) \quad}{Ensure ${\bf c}_{j\rightarrow{i}}$ and ${\bf c}_{j}$ infer \textit{different} task identities} + \quad m\bigg]_{+},
\end{equation}

\begin{figure}[!t]
\begin{minipage}{0.5\textwidth}
\begin{algorithm}[H]
    \caption{Distillation and triplet loss}\label{algo_mtbrl}
    {\bf Input}: Batches $\{\mathcal{B}_{i}\}_{i=1}^K$; BCQ-trained $\{Q_i\}^K_{i=1}$, $\{G_i\}^K_{i=1}$, and $\{\xi_i\}^K_{i=1}$; randomly initialized $Q_D$, $G_D$ and $\xi_D$ jointly parameterized by $\theta$; task inference module $q_\phi$ with randomly initialized $\phi$
    \begin{algorithmic}[1]
        \Repeat
            \State Sample context set ${\bf c}_i$ from $\mathcal{B}_{i}, \forall i$
            \State Obtain relabelled transitions ${\bf c}_{j\rightarrow i}$ according to Eq. \ref{eq_relabel} for all pair of task $i, j$
            \State Calculate $\mathcal{L}_{triplet}$ using Eq. \ref{eq_triplet_all_task}
        \State Calculate $\mathcal{L}_Q, \mathcal{L}_G, \mathcal{L}_\xi$ using Eq. \ref{eq_loss_qi}, \ref{eq_loss_gi}, \ref{eq_loss_xi_i}
        \State Calculate $\mathcal{L}$ using Eq. \ref{eq_loss_final}
        \State Update $\theta, \phi$ to minimize $\mathcal{L}$
        \Until{Done}
    \end{algorithmic}
\end{algorithm}
\end{minipage}
\hfill
\begin{minipage}{0.45\textwidth}
\begin{figure}[H]
\centering
\includegraphics[width=0.6\textwidth]{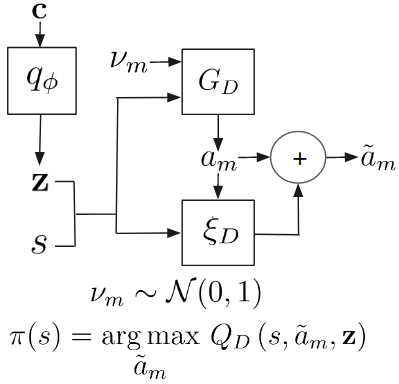}
\caption{Action selection. Given $s$, $G_D$ generates candidate actions $a_m$. 
$\xi_D$ generates small corrections for the actions $a_m$. The policy takes the corrected action $\tilde{a}_m$ with the highest value as estimated by $Q_D$.}
\label{fig:action}
\end{figure}
\end{minipage}
\end{figure}

where $m$ is the triplet margin, $[\cdot]_{+}$ is the ReLU function and $d$ is a divergence measure. $q_\phi$ outputs the posterior over task identity, we thus choose $d$ to be the KL divergence.

Minimizing Eq. \ref{eq_triplet_task_i} accomplishes two goals. It encourages the task inference module $q_\phi$ to infer similar task identities when given either ${\bf c}_i$ or ${\bf c}_{j\rightarrow{i}}$ as input. It also encourages $q_\phi$ to infer different task identities for ${\bf c}_j$ and ${\bf c}_{j\rightarrow{i}}$. We emphasize that the task inference module can not learn to correlate \textit{only} the state-action pairs with the task identity since ${\bf c}_j$ and ${\bf c}_{j\rightarrow{i}}$ contain the same state-action pairs, but they correspond to different task identities. To minimize Eq. \ref{eq_triplet_task_i}, the module must model the correct conditional dependency $P(Z|S, A, R)$ when inferring the task identity. 

Eq. \ref{eq_triplet_task_i} calculates the triplet loss when we use the learned reward function of task $i$ to relabel transitions from the remaining tasks. Following similar procedures for the remaining tasks lead to the loss:
\begin{equation}\label{eq_triplet_all_task}
    \mathcal{L}_{triplet} =  \frac{1}{K} \sum_{i=1}^{K} \mathcal{L}^i_{triplet}.
\end{equation}
The final loss to train the randomly initialized task inference module $q_\phi$, the distilled value functions $Q_D$, the distilled candidate action generator $G_D$, and the distilled perturbation generator $\xi_D$ is:
\begin{equation}\label{eq_loss_final}
    \mathcal{L} = \mathcal{L}_{triplet} + \mathcal{L}_Q + \mathcal{L}_G + \mathcal{L}_\xi.
\end{equation}
Alg. \ref{algo_mtbrl} illustrates the pseudo-code for the second phase of the distillation procedure. Detailed pseudo-code of the two-phases distillation procedures can be found in \autoref{sec_detailed_distill}. Fig. \ref{fig:action} briefly describes action selection from the multi-task policy. \autoref{sec_action} provides detailed explanations. In theory, we can also use the relabelled transitions in Eq. \ref{eq_relabel} to train the single-task BCQ policy in the first phase, which we do not since we focus on task inference in this work.

\section{Discussions}

The issue of learning the wrong dependency does not surface when multi-task policies are tested in Atari tasks because their state space do not overlap \cite{ActorMimicParisotto2015,hessel2019multi,impala2018}.
Each Atari task has distinctive image-based state. The policy can perform well even when it only learns to correlate the state to the task identity. When Mujoco tasks are used to test online multi-task algorithms \cite{zintgraf2020varibad, fakoor2019meta}, the wrong dependency becomes self-correcting. If the policy infers the wrong task identity, it will collect training data which increases the overlap between the datasets of the different training tasks, correcting the issue overtime. However, in the batch setting, the policy can not collect more transitions to self-correct inaccurate task inference. Our insight also leads to exciting possibility to incorporate mechanism to quickly infer the correct causal relationship and improve sample efficiency in Multi-task RL, similar to how causal inference method has motivated new innovations in imitation learning \cite{de2019causal}. 

Our first limitation is the reliance on the generalizability of simple feedforward NN. Future research can explore more sophisticated architecture, such as Graph NN with reasoning inductive bias \cite{xu2019can,scarselli2008graph,wu2020comprehensive,zhou2018graph} or structural causal model \cite{pearl2010introduction, pearl2009causal}, to ensure accurate task inference. We also assume the learnt reward function of one task can generalize to state-action pairs from the other tasks, even when their state-action visitation frequencies do not overlap significantly. To increase the prediction accuracy, we use a reward ensemble to estimate epistemic uncertainty (\autoref{sec_ensemble}). We note that the learnt reward functions do not need to generalize to every state-action pairs, but only enough pairs so that the task inference module is forced to consider the rewards when trained to minimize Eq. \ref{eq_triplet_task_i}. Crucially, we do not need to solve the task inference challenge while learning the reward functions and using them for relabelling, allowing us to side-step the challenge of task inference.

The second limitation is in scope. We only demonstrate our results on tasks using proprioceptive states. Even though they represent high-dimensional variables in a highly nonlinear ODE, the model does not need to tackle visual complexity. The tasks we consider also have relatively dense reward functions and not binary reward functions. These tasks, such as navigation and running, are also quite simple in the spectrum of possible tasks we want an embodied agents to perform. These limitations represent exciting directions for future work.

Another interesting future direction is to apply supervised learning self-distillation techniques \cite{xie2019selftraining, mobahi2020selfdistillation}, proven to improve generalization, to further improve the distillation procedure. To address the multi-task learning problem for long-horizon tasks, it would also be beneficial to consider skill discovery and composition from the batch data \cite{peng2019MCP, sharma2020emergent}. However, in this setting, we still need effective methods to infer the correct task identity to perform well in unseen tasks. Our explanation in Sec. \ref{sec_algo} only applies when the tasks differ in reward function. Extending our approach to task distributions with varying transition functions is trivial. Sec. \ref{sec_exp} provide experimental results for both cases.

\section{Experiment Results}\label{sec_exp}
We demonstrate the performance of our proposed algorithm (Sec. \ref{sec_exp_eval}) and ablate the different design choices (Sec. \ref{sec_exp_ablation}). Sec. \ref{sec_exp_init_sac} shows that the multi-task policy can serve as a good initialization, significantly speeding up training on unseen tasks. \autoref{sec_hyper_param} provides all hyper-parameters.

\subsection{Performance evaluation on unseen tasks}\label{sec_exp_eval}


\begin{figure}[!t]
\begin{minipage}{0.28\textwidth}
\begin{figure}[H]
\centering
\includegraphics[width=\textwidth]{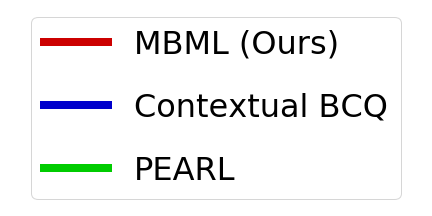}
\caption{Results on unseen test tasks. 
x-axis is training epochs. 
y-axis is average episode returns.
The shaded areas denote one std.
  }\label{fig:testing_performance}
\end{figure}
\end{minipage}
\hfill
\addtocounter{figure}{-1}
\begin{minipage}{0.69\textwidth}

  \makebox[\textwidth]{
  
  \begin{subfigure}{\mujocobaselinefigsize\paperwidth}
    \includegraphics[width=\linewidth]{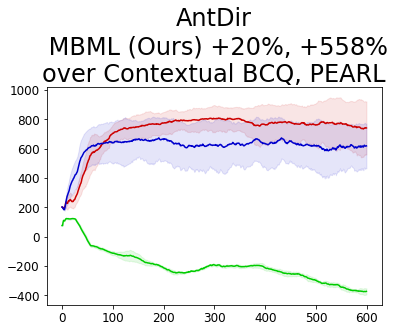}
  \end{subfigure}
  
  \begin{subfigure}{\mujocobaselinefigsize\paperwidth}
    \includegraphics[width=\linewidth]{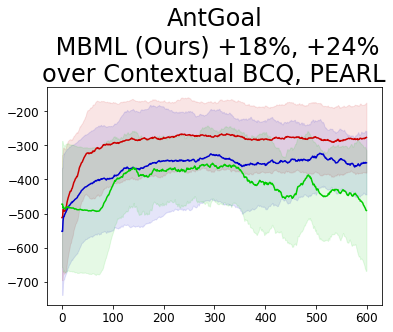}
  \end{subfigure}
    
  \begin{subfigure}{\mujocobaselinefigsize\paperwidth}
    \includegraphics[width=\linewidth]{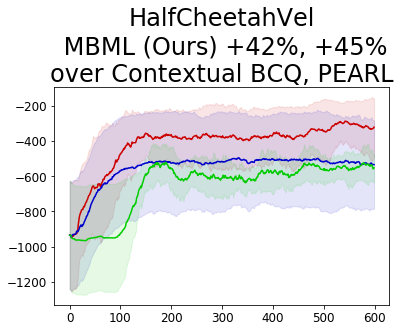}
  \end{subfigure}}
  
  \makebox[\textwidth]{
  \begin{subfigure}{\mujocobaselinefigsize\paperwidth}
    \includegraphics[width=\linewidth]{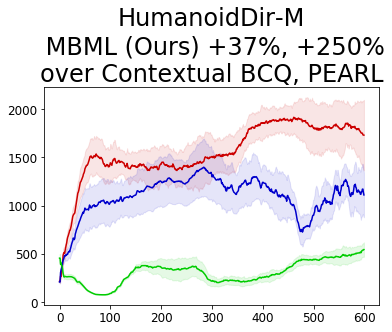}
  \end{subfigure}
  
  \begin{subfigure}{\mujocobaselinefigsize\paperwidth}
    \includegraphics[width=\linewidth]{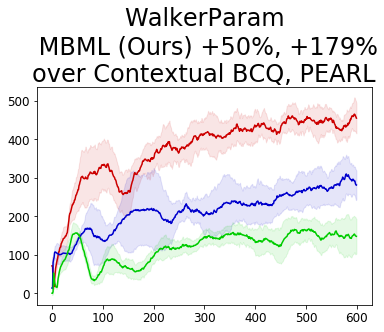}
  \end{subfigure}
  
  \begin{subfigure}{\mujocobaselinefigsize\paperwidth}
    \includegraphics[width=\linewidth]{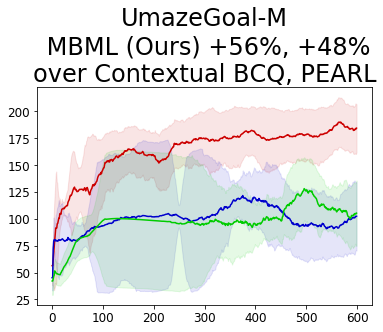}
  \end{subfigure}}

\end{minipage}
\end{figure}

We evaluate in five challenging task distributions from MuJoCo \cite{todorov2012mujoco} and a modified task distribution UmazeGoal-M from D4RL \cite{fu2020d4rl}. In AntDir and HumanoidDir-M, a target direction defines a task. The agent maximizes returns by running with maximal speed in the target direction. In AntGoal and UmazeGoal-M, a task is defined by a goal location, to which the agent should navigate. In HalfCheetahVel, a task is defined as a constant velocity the agent should achieve. We also consider the WalkerParam environment where random physical parameters parameterize the agent, inducing different transition functions in each task. The state for each task distribution is the OpenAI gym state. We do not include the task-specific information, such as the goal location or the target velocity in the state. The target directions and goals are sampled from a $120^{\circ}$ circular arc. Details of these task distributions can be found in Appendix \ref{sec_env_detail}.

We argue that the version of HumanoidDir used in prior works does not represent a meaningful task distribution, where a single task policy can already achieve the optimal performance on unseen tasks. We thus modify the task distribution so that a policy has to infer the task identity to perform well, and denote it as HumanoidDir-M. More details of this task distribution can be found in \autoref{sec_humanoid}.

There are two natural baselines. The first is by modifying PEARL \cite{rakelly2019efficient} to train from the batch, instead of allowing PEARL to collect more transitions. We thus do not execute line $1-10$ in Algorithm 1 in the PEARL paper. On line 13, we sample the context and the RL batch uniformly from the batch. The second baseline is Contextual BCQ. We modify the networks in BCQ to accept the inferred task identity as input. We train the task inference module using the gradient of the value function loss. MBML and the baselines have the same network architecture. We are very much inspired by PEARL and BCQ. However, we do not expect PEARL to perform well in our setting because it does not explicitly handle the difficulties of learning from a batch without interactions. We also expect that our proposed algorithm will outperform Contextual BCQ thanks to more robust task inference. 

We measure performance by the average returns over unseen tasks, sampled from the same task distribution. We do not count the first two episodes' returns \cite{rakelly2019efficient}. We obtain the batch for each training task by training Soft Actor Critic (SAC) \cite{haarnoja2018soft} with a fixed number of environment interactions. \autoref{sec_env_baseline} provide more details on the environment setups and training procedures of the baselines. 

\begin{wrapfigure}{R}{0.259\textwidth}
\vspace{-0.6em}
\flushright
\begin{subfigure}{0.155\paperwidth}
    \includegraphics[width=\linewidth, right]{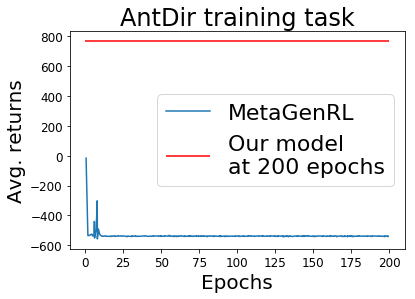}
\end{subfigure}
\caption{MetaGenRL quickly diverges and does not recover.} 
\label{fig:metagenrl}  
\vspace{0.05em}
\end{wrapfigure}

From Fig. \ref{fig:testing_performance}, MBML outperforms the baselines by a healthy margin in all task distributions. Even though PEARL does not explicitly handle the challenge of training from an offline batch, it is remarkably stable, only diverging in AntDir. Contextual BCQ is stable, but converges to a lower performance than MBML in all task distributions. An astude reader will notice the issue of overfitting, for example Contextual BCQ in HumanoidDir-M. Since our paper is not about determining early stopping conditions and to ensure fair comparisons among the different algorithms, we compute the performance comparisons using the best results achieved by each algorithm during training.

We also compare with MetaGenRL \cite{metagenrl}. Since it relies on DDPG \cite{ddpg} to estimate value functions, which diverges in Batch RL \cite{fujimoto2019off}, we do not expect it to perform well in our setting. Fig. \ref{fig:metagenrl} confirms this, where its performance quickly plummets and does not recover with more training. 
Combining MetaGenRL and MBML is interesting since MetaGenRL generalizes to out-of-distribution tasks.

\subsection{Ablations}\label{sec_exp_ablation}

\begin{figure}[!t]
\begin{minipage}{0.28\textwidth}
\begin{figure}[H]
\centering
\includegraphics[width=\textwidth]{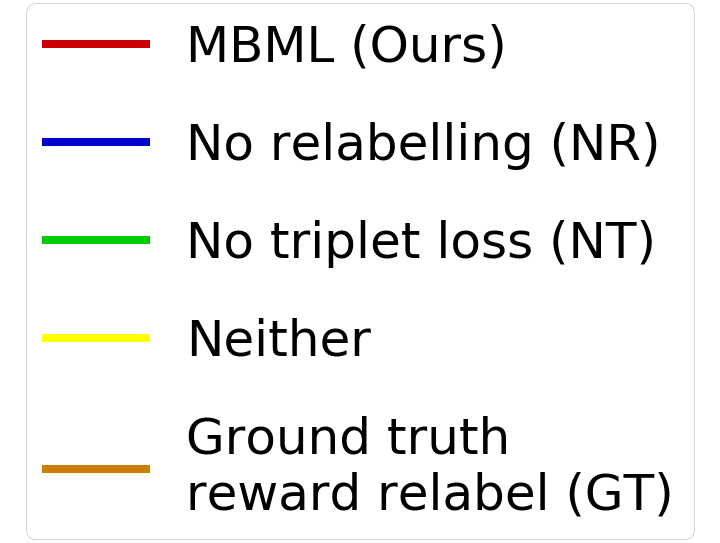}
\caption{Ablation study. 
x-axis is training epochs. 
y-axis is average episode returns.
The shaded areas denote one std.
}\label{fig:ablation}
\end{figure}
\end{minipage}
\hfill
\addtocounter{figure}{-1}
\begin{minipage}{0.70\textwidth}

  \makebox[\textwidth]{
  
  \begin{subfigure}{\mujocobaselinefigsize\paperwidth}
    \includegraphics[width=\linewidth]{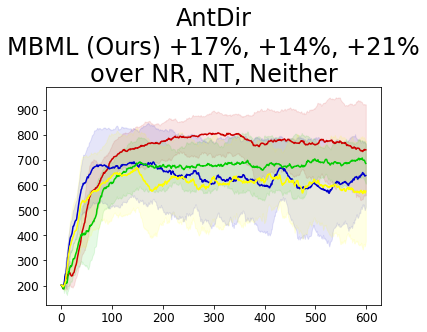}
  \end{subfigure}
  
  \begin{subfigure}{\mujocobaselinefigsize\paperwidth}
    \includegraphics[width=\linewidth]{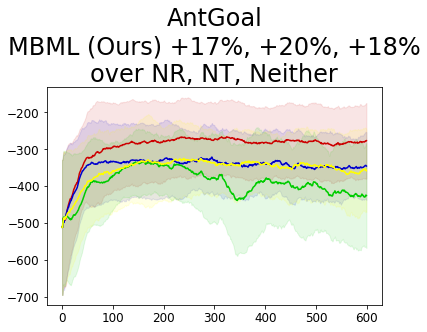}
  \end{subfigure}
    
  \begin{subfigure}{\mujocobaselinefigsize\paperwidth}
    \includegraphics[width=\linewidth]{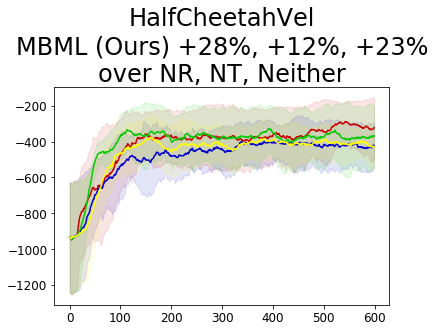}
  \end{subfigure}}
  
  \makebox[\textwidth]{
  \begin{subfigure}{\mujocobaselinefigsize\paperwidth}
    \includegraphics[width=\linewidth]{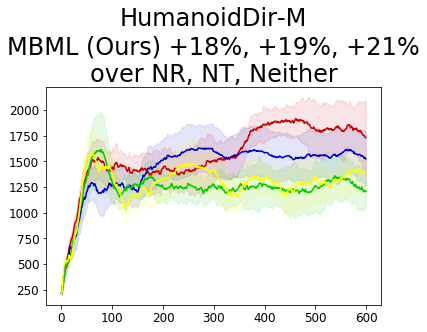}
  \end{subfigure}
  
  \begin{subfigure}{\mujocobaselinefigsize\paperwidth}
    \includegraphics[width=\linewidth]{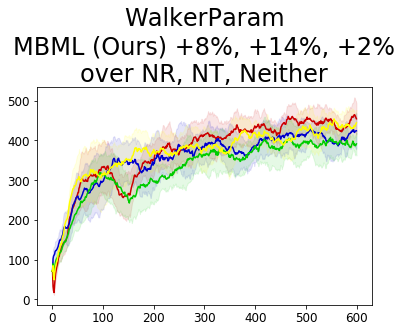}
  \end{subfigure}
  
  \begin{subfigure}{\mujocobaselinefigsize\paperwidth}
    \includegraphics[width=\linewidth]{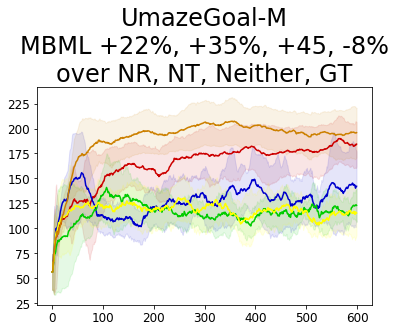}
  \end{subfigure}}

\end{minipage}

\end{figure}

We emphasize that our contributions lie in the triplet loss design coupled with transitions relabelling. Below, we provide ablation studies to demonstrate that both are crucial to obtain superior performance.

\textbf{No relabelling.} To obtain  hard negative examples, we search over a mini-batch to find the hardest positive-anchor and negative-anchor pairs, a successful and strong baseline from metric learning \cite{hermans2017defense}. This requires sampling $N$ context sets $\{{\bf c}^n_i\}_{n=1}^N$ for each task $i$, where $n$ indexes the context sets sampled for each task. 
Let $K$ be the number of training tasks, the triplet loss is:
\begin{equation}\label{eq_reform_triplet}
    \frac{1}{K} \sum_{i=1 }^{K} \bigg[\max_{n,n'=1,\ldots, N} \quad d \left(q_{\phi}({\bf c}^n_{i}\big), q_{\phi} ({\bf c}^{n'}_{i}) \right)
    -  \min_{\substack{n,n'=1,\ldots, N \\ j=1,\ldots,K, j \neq i}} \quad d \left(q_{\phi}({\bf c}^n_{i}\big), q_{\phi} ({\bf c}^{n'}_{j}) \right) + m\bigg]_{+}.
\end{equation}
The $max$ term 
finds the positive-anchor pair
for task $i$ 
by considering every pair of context sets from task $i$ and selecting the pair with the largest divergence in the posterior over task identities. The $min$ term
finds the negative-anchor pair 
for task $i$ 
by considering every possible pair between the context sets sampled for task $i$ and the context sets sampled for the other tasks. It then selects the pair with the lowest divergence in the posterior over task identities
as the negative-anchor pair.

\textbf{No triplet loss.} We train the task inference module using only gradient of the value function distillation loss (Eq. \ref{eq_loss_qi}).
To use the relabelled transitions, the module also takes as input the relabelled transitions during training.
More concretely, given the context set ${\bf c}_i$ from task $i$, we sample an equal number of relabelled transitions from the other tasks $\tilde{{\bf c}}_i \sim \cup_j {{\bf c}_{j \rightarrow{i} } }$. During training, the input to the task inference module is the union of the context set ${\bf c}_i$ and the sampled relabelled transitions $\tilde{{\bf c}}_i$. In the full model, we also perform similar modification to the input of the module during training.

\textbf{No transition relabelling and no triplet loss.} This method is a simple combination of a task inference module and the distillation process. We refer to this algorithm as \textbf{Neither} in the graphs.

Fig. \ref{fig:ablation} compares our full model and the ablated versions. Our full model obtains higher returns than most of the ablated versions. For WalkerParam, our full model does not exhibit improvement over  \textbf{Neither}. However, from Fig. \ref{fig:testing_performance}, our full model significantly outperforms the baselines. We thus conclude that, in WalkerParam, the improvement over the baselines comes from distillation.

\begin{wrapfigure}{R}{0.239\textwidth}
\flushright
\vspace{-0.5em}
\begin{subfigure}{0.16\paperwidth}
    \includegraphics[width=\linewidth]{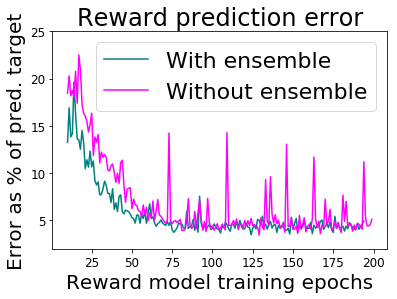}
\end{subfigure}
\caption{Error on unseen task.}\label{fig:reward_error} 
\vspace{-0.2em}
\end{wrapfigure}

Comparing to the \textbf{No triplet loss} ablation, transition relabelling leads to more efficient computation of the triplet loss. Without the relabelled transitions, computing Eq. \ref{eq_reform_triplet} requires $O(K^2 N^2)$. Our loss in Eq. \ref{eq_triplet_all_task} only requires $O(K^2)$. We also need to relabel the transitions only once before training the multi-task policy. It is also trivial to parallelize across tasks. 

We also study reward estimation accuracy. Fig. \ref{fig:reward_error} shows that our reward model achieves low error on state-action pairs from another task, both with and without an ensemble. We also compare MBML against an ablated version that uses the ground truth reward function for relabelling on UmazeGoal-M. The model trained using the ground truth reward function only performs slightly better than the model trained using the learned reward function. We include in \autoref{sec_ablation} experiments on margin sensitivity analysis and the benefit of the reward ensemble.

\subsection{Using the multi-task policy to enable faster convergence when training on unseen tasks}\label{sec_exp_init_sac}

While the multi-task policy generalize to unseen tasks, its performance is not optimal. If we allow further training, 
initializing networks with our multi-task policy significantly speeds up convergence to the optimal performance. 

The initialization process is as followed. Given a new task, we use the multi-task policy to collect 10K transitions. 
We then train a new policy to imitate the actions taken by maximizing their log likelihood. As commonly done, the new policy outputs the mean and variance of a diagonal Gaussian distribution. The new policy does not take a task identity as input. The task inference module infers a task identity {\bf z} from the 10K transitions. Fixing {\bf z} as input, the distilled value function $Q_D$ initializes the new value function. Given the new policy and the initialized value function, 
we train them with SAC by collecting more data.
To stabilize training, we perform target policy smoothing \cite{TD3} and double-Q learning \cite{van2016deep} by training two identically initialized value functions with different mini-batches (pseudo-codes and more motivations in Appendix \ref{sec_sac_our_methods}).

Fig. \ref{fig:SAC_init} compares the performance of the policies initialized with our multi-task policy to randomly initialized policies. 
Initializing the policies with the MBML policy significantly increases convergence speed in all five task distributions, demonstrating our method's robustness.
Even in the complex HumanoidDir-M task distribution, our method significantly speeds up the convergence, requiring only 85K environment interactions, while the randomly initialized policies require 350K, representing a $76\%$ improvement in sample efficiency.
Similar conclusions hold when comparing against randomly initialized SAC where the two value functions are trained using different mini-batches (Appendix \ref{sec_init_counterpart}). 
We also note that our initialization method does not require extensive hyper-parameter tuning.


\begin{figure*}[t]
    \centering
  \makebox[\textwidth]{
  \begin{subfigure}{0.125\paperwidth}
    \includegraphics[width=\linewidth]{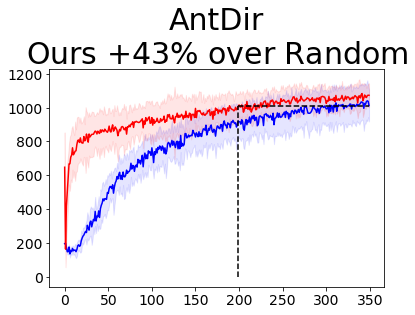}
  \end{subfigure}
  
  \begin{subfigure}{0.125\paperwidth}
    \includegraphics[width=\linewidth]{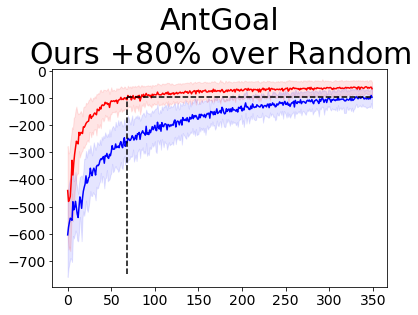}
  \end{subfigure}
    
  \begin{subfigure}{0.125\paperwidth}
    \includegraphics[width=\linewidth]{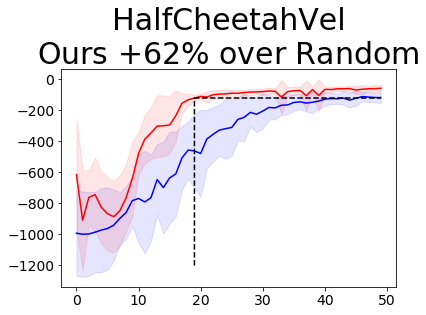}
  \end{subfigure}
  
  \begin{subfigure}{0.125\paperwidth}
    \includegraphics[width=\linewidth]{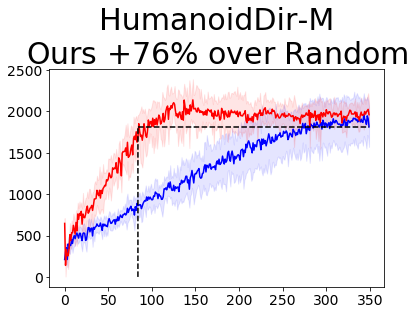}
  \end{subfigure}
  
  \begin{subfigure}{0.125\paperwidth}
    \includegraphics[width=\linewidth]{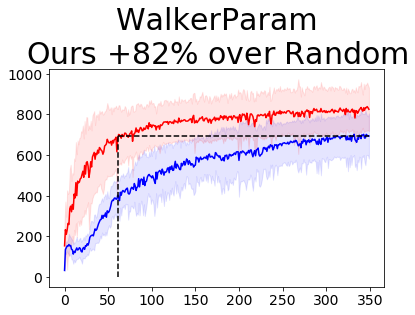}
  \end{subfigure}}
  \small{\color{red}--- }: SAC initialized by our multi-task policy (Ours) \qquad  {\color{blue}--- }: Randomly initialized SAC (Random)
  \caption{Initialization results. 
  x-axis is number of interactions in thousands. 
  y-axis is the average episode returns over unseen tasks.
  The shaded areas denote one std. }
\label{fig:SAC_init}
\end{figure*}
\section{Related Works}\label{sec_related_works}

\textbf{Batch RL} 
Recent advances in Batch RL \cite{agarwal2019optimistic,kumar2019stabilizing,fujimoto2019off,chen2019bail,kumar2020conservative} focus on the single-task setting, which does not require training a task inference module. Thus they are not directly applicable to the Multi-task Batch RL. \cite{siegel2020keep, cabi2020ScalingDR} also consider the multi-task setting but assume access to the ground truth task identity and reward function of the test tasks. Our problem setting also differs, where the different training batches do not have significant overlap in state-action visitation frequencies, leading to the challenge of learning a robust task inference module.

\textbf{Task inference in multi-task setting}
The challenge of task inference in a multi-task setting has been tackled under 
various umbrellas. Meta RL \cite{rakelly2019efficient, zintgraf2020varibad,fakoor2019meta,humplik2019meta,lan2019meta,saemundsson2018meta,CAVIA} trains a task inference module to infer the task identity from a context set.
We also follow this paradigm. However, our setting presents additional challenge to train a robust task inference module, which motivates our novel 
triplet loss design. 
As the choice of loss function is crucial to train an successful task inference module in our settings, we will explore the other loss functions, e.g. loss functions discussed in \cite{roth2020revisiting}, in future work.
Other multi-task RL works \cite{espeholt2018impala, yang2020multi, yumulti,d2019sharing} focus on training a good multi-task policy, rather than the task inference module, which is an orthogonal research direction to ours.

\textbf{Meta RL} Meta RL \cite{lan2019meta,LearnToReinforceLearnWang2016,duan2016rl,finn2017maml,nichol2018Reptile,houthooft2018evolvedpg} optimizes for quick adaptation. However, they require interactions with the environment during training. Even though we do not explicitly optimize for quick adaptation, we demonstrate that initializing a model-free RL algorithm with our policy significantly speeds up
convergence on unseen tasks. \cite{fakoor2019meta} uses the data from the training tasks to speed up convergence when learning on new tasks by propensity estimation techniques. This approach is orthogonal to ours and can potentially be combined to yield even greater performance improvement.



\clearpage

\section*{Acknowledgement}

We thank Fangchen Liu (UC Berkeley) for pointing out a figure issue right before the paper submission deadline. We thank Chaochao Lu (University of Cambridge) for introducing us to causality. Computing needs were supported by the Nautilus Pacific Research Platform.

\section*{Broader Impact}

\subsection*{Positive impact}

Our work provides a solution to learn a policy that generalizes to a set of similar tasks from only observational data. The techniques we propose have great potential to benefit various areas of the whole society. For example in the field of healthcare, we hope the proposed triplet loss design with hard negative mining can enable us to robustly train an automatic medical prescription system from a large batch of medical histories of different diseases and further generalize to new diseases \cite{choi2019meta}, e.g., COVID-19.  Moreover, in the field of robotics, our methods can enable the learning of a single policy that solves a set of similar unseen tasks from only historical robot experiences, which tackles the sample efficiency issues given that sampling is expensive in the field of real-world robotics \cite{cabi2020ScalingDR}. Even though in some fields that require safe action selections, e.g, autonomous driving \cite{geiger2012we} and medical prescription, our learned policy cannot be immediately applied, it can still serve as a good prior to accelerate further training.  

\subsection*{Negative impact}

Evidently, the algorithm we proposed is a data-driven methods. Therefore, it is very likely that it will be biased by the training data. Therefore, if the testing tasks are very different from the training tasks, the learned policy may even result in worse behaviors than random policy, leading to safety issues. This will motivate research into safe action selection and distributional shift identification when learning policies for sequential process from only observational data.

\clearpage

\bibliographystyle{unsrt}
\bibliography{bib_mtbrl}

\clearpage

\appendix
\appendixpage
\addappheadtotoc

\section{Symbol definition}\label{sec_def_var}

\begin{table}[H]
\small
\centering
\caption{Symbol definition. Some of the symbol are overloaded. We make sure each term is clearly defined given the context.}\label{table:def_variables}
\begin{tabular}{ |l|l|l| }
        \hline
        {\bf Symbol} & {\bf Definition} & {\bf Dimension}\\ 
        \hline
        $\mathcal{S}$  & state space & $\mathbb{R}^{N_s}$\\
        
        $\mathcal{A}$ & action space & $\mathbb{R}^{N_a}$\\
        
        $T$  & transition function & $\mathbb{R}^{N_s + N_a}\rightarrow \mathbb{R}^{N_s}$\\ 

        $T_0$ & initial state distribution & $\mathbb{R}^{N_s}\rightarrow [0, 1]$\\
        
        $R$ & reward function & $\mathbb{R}^{N_s + N_a + N_s}\rightarrow \mathbb{R}$\\

        $H$ & horizon & $\mathbb{N}_{+}$\\
        
        $M$ & MDP $M = (\mathcal{S}, \mathcal{A}, T, T_0, R, H)$, which defines a task & -\\
        
        $p(M)$ & task distribution & -\\
        
        $N_\theta$ & dimension of the policy parameter & $\mathbb{N}_{+}$\\
        
        $K$ & number of task & $\mathbb{N}_{+}$\\
        
        $\tau_M$ & trajectory generated by interacting with $M$ & - \\
        $s_t$ & state at time step $t$ & $\mathbb{R}^{N_s}$\\
        
        $a_t$ & action selected at time step $t$ & $\mathbb{R}^{N_a}$\\
        
        $\tilde{a}$ & corrected action & $\mathbb{R}^{N_a}$\\
        
        $s'_t$ & state at time step $t$ & $\mathbb{R}^{N_s}$\\
        
        $\pi_\theta$ & policy function,  parameterized by $\theta$ & $\mathbb{R}^{N_s}\rightarrow \mathbb{R}^{N_a}$\\
        
        $N$ & number of transition tuples from one task batch & $\mathbb{N}_{+}$\\
        
        $\mathcal{B}_i$ & batch of transition tuples for task $i$ & -\\
        
        $\hat{R}_i$ & learned reward function for task $i$ & $\mathbb{R}^{N_s + N_a + N_s}\rightarrow \mathbb{R}$\\
        
        $N_c$ & number of transition tuples in a context set & $\mathbb{N}_{+}$\\
        
        ${\bf c}_i$  & context set for task $i$ & $\mathbb{R}^{(N_s + N_a + N_s) * N_c}$\\
        
        ${\bf c}_{j\rightarrow{i}}$  & relabeled context set by uing $\hat{R}_i$ & $\mathbb{R}^{(N_s + N_a + N_s) * N_c}$\\
        
        $\tilde{\bf c}_i$  & Union of relabeled context set by uing $\hat{R}_i$ & $\mathbb{R}^{(N_s + N_a + N_s) * N_c * (K - 1)}$\\
        
        ${\bf z}_i$  & task identity for task $i$ & $\mathbb{R}^{N_z}$ \\
        
        $q_\phi$ & task inference module, parameterized by $\phi$ & $\mathbb{R}^{(N_s + N_a + N_s) * N_c}\rightarrow ({N_z}, {N_z})$\\
        
        $S$, $A$, $R$, $Z$ & random variables: states, actions, rewards, task identity & - \\
        
        $J_{M_i}(\pi_\theta)$ & expected sum of rewards in$ M_i$ induced by policy $\pi_\theta $ & $\mathbb{R}^{N_\theta}\rightarrow \mathbb{R}$\\
        
        $J(\pi_\theta)$ & expected sum of rewards in $p(M)$ induced by policy $\pi_\theta $ & $\mathbb{R}^{N_\theta}\rightarrow \mathbb{R}$\\
        
        $Q$ & Q value function &  $\mathbb{R}^{N_s + N_a}\rightarrow \mathbb{R}$\\
        
        $Q_i$ & Q value function for task $i$ &  $\mathbb{R}^{N_s + N_a}\rightarrow \mathbb{R}$\\
        
        $Q_D$ & distilled Q value function &  $\mathbb{R}^{N_s + N_a + N_z}\rightarrow \mathbb{R}$\\
        
        $G$ & candidate action generator & $\mathbb{R}^{N_s + 1}\rightarrow \mathbb{R}^{N_a}$\\
        
        $G_i$ & candidate action generator for task $i$ & $\mathbb{R}^{N_s + 1}\rightarrow \mathbb{R}^{N_a}$\\
        
        $G_D$ & distilled candidate action generator &  $\mathbb{R}^{N_s + 1 + N_z}\rightarrow \mathbb{R}^{N_a}$\\
        
        $\xi$ & perturbation generator & $\mathbb{R}^{N_s + N_a}\rightarrow \mathbb{R}^{N_a}$\\
        
        $\xi_i$ & perturbation generator for task $i$ & $\mathbb{R}^{N_s + N_a}\rightarrow \mathbb{R}^{N_a}$\\
        
        $\xi_D$ & distilled perturbation generator &  $\mathbb{R}^{N_s + N_a + N_z}\rightarrow \mathbb{R}^{N_a}$\\
        
        $\mathcal{N}(0, 1)$ & standard Gaussian distribution & -\\
        
        $\nu$ & noise sampled from standard Gaussian distribution & $\mathbb{R}$\\
        
        $m$ & triplet margin & $\mathbb{R}$\\
        
        $d(\cdot)$ & divergence measure & -\\
        
        KL & KL divergence & -\\
        
        $\bar{\cdot}$ & stop gradient operation & -\\
        
        $\mathcal{L}_{Q}$ & loss function to distill $Q_D$ & -\\
        
        $\mathcal{L}_{G}$ & loss function to distill $G_D$ & -\\
        
        $\mathcal{L}_{\xi}$ & loss function to distill $\xi_D$ & -\\
        
        $\mathcal{L}_{distill}$ & total distillation loss & -\\
        
        $\mathcal{L}^i_{triplet}$ & triplet loss for task $i$ & -\\
        
        $\mathcal{L}_{triplet}$ & mean triplet loss across all tasks & -\\
        
        $\mathcal{L}$ & final loss: $\mathcal{L} = \mathcal{L}_{triplet} + \mathcal{L}_{distill}$ & -\\
         
        \hline
        \end{tabular}
\end{table}

\clearpage

\section{Action selection of BCQ policy}\label{sec_action_bcq}
\begin{figure}[H]
    \centering
    \includegraphics[width=\textwidth]{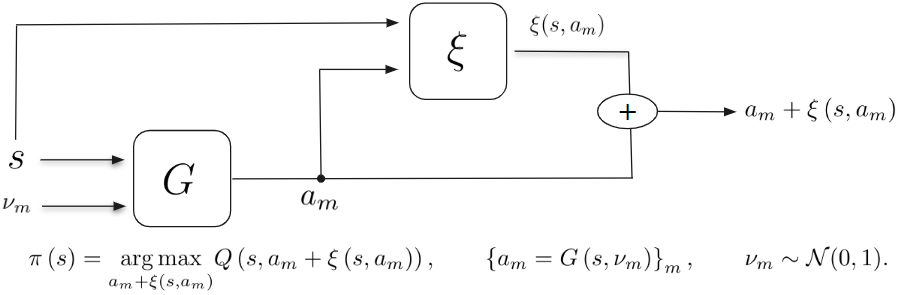}
    \caption{Action selection procedure of BCQ.}
    \label{fig:bcq-action-selection}
\end{figure}

In this section, we provide the detailed action selection procedures for BCQ. To pick action given a state $s$, we first sample a set of small noises $\{\nu_m\}_m$ from the standard Gaussian distribution. For each $\nu_m$, the candidate action generator $G$ will generate a candidate action $a_m$ for state $s$. For each of the candidate actions $a_m$, the perturbation model $\xi$ will generate a small correction term $\xi(s, a_m)$ by taking as input the state-candidate action pair. Therefore, a set of corrected candidate actions $\{a_m + \xi(s, a_m)\}_{m}$ will be generated for the state $s$. The corrected candidate action with the highest estimated $Q$ value will be selected as $\pi\left(s\right)$.

\clearpage

\section{Hyper-parameters}\label{sec_hyper_param}

\subsection{Hyper-parameters of our proposed models}\label{sec_hp_distillation}

\begin{table}[H]
\small
\centering
\caption{Hyper-parameters of our proposed model}\label{table:hp-distillation}
\begin{tabular}{ |l|l| }
        \hline
        {\bf Hyper-parameters} & {\bf Value} \\ 
        \hline
        Number of evaluation episodes  & 5 \\ 

        Task identity dimension  & 20 \\ 
        
        Number of candidate actions & 10 \\

        Learning rate & 0.0003 \\
        
        Training batch size & 128  \\ 
        
        Context set size & 64 \\
        
        KL regularization weighting term $\beta$ & 0.1 \\
        
        Triplet margin $m$  & 2.0 \\ 
        
        \hline
        Reward prediction ensemble $\sigma_{\text{threshold}}$ & AntDir, AntGoal: 0.1\\ & WalkerParam: 0.1 \\ & HumanoidDir-M: 0.2\\  & HalfCheetahVel: 0.05\\ & UmazeGoal-M: 0.02 \\
        \hline
        Next state prediction ensemble $\sigma_{\text{threshold}}$ & 0.1 \\
        \hline
        $Q_D$ architecture & MLP with 9 hidden layers, 1024 nodes each, ReLU activation \\ 

        $G_D$ architecture & MLP with 7 hidden layers, 1024 nodes each, ReLU activation \\ 

        $\xi_D$ architecture & MLP with 8 hidden layers, 1024 nodes each, ReLU activation \\
        \hline
        \end{tabular}
\end{table}

\begin{table}[H]
\small
\centering
    \caption{Hyper-parameters of reward and next state prediction ensemble}\label{table:hp-ensemble}
    \begin{tabular}{ |l|l| }
            \hline
            {\bf Hyper-parameters} & {\bf Value} \\ 
            \hline
            Learning rate & 0.0003 \\
            
            Training batch size & 128 \\
            
            Reward prediction ensemble size & 20 \\
            
            Reward prediction network architecture & MLP with 1 hidden layers, 128 nodes, ReLU activation \\ 
            
            Next state prediction ensemble size & 20 \\
            
            Next state prediction network architecture & MLP with 6 hidden layers, 256 nodes each, ReLU activation \\ 
            \hline
    \end{tabular}
\end{table}

\autoref{table:hp-distillation} provides the hyper-parameters for our proposed model and all of its ablated versions (Sec. \ref{sec_algo}, Sec. \ref{sec_exp_ablation}). The hyper-parameters for the reward ensembles and next state prediction ensembles are provided in \autoref{table:hp-ensemble}. Our model uses the task inference module from PEARL with the same architecture, described in \autoref{table:hp-pearl}. Since the scale of the reward in different task distributions are different, we need to use different values for the reward prediction ensemble threshold $\sigma_{\text{threshold}}$.

We did not conduct extensive search to determine the hyper-parameters. Instead, we reuse some default hyper-parameter settings from the other multi-task learning literature on the MuJoCo benchmarks \cite{rakelly2019efficient,fakoor2019meta}. As for the architecture of the distillation networks, we select reasonably deep networks.

When using BCQ to train the single-task policies in the first phase of the distillation procedure, we use the default hyper-parameters in the official implementation of BCQ, except for the learning rate, which decreases from $0.001$ to $0.0003$. We find lowering the learning rate leads to more stable learning for BCQ.

\subsection{Hyper-parameters of Contextual BCQ}

For Contextual BCQ, the value function, decoder, and perturbation model have the same architecture as $Q_D, G_D, \xi_D$ in our model. The encoder also has the same architecture as the decoder. The task inference module has the same architecture as the task inference module in PEARL, described in \autoref{table:hp-pearl}. 

The context set size used during training Contextual BCQ is $128$, twice the size of the context set in our model. This is because during training of our model, we use the combination of context transitions and the same number of relabelled transitions from the other tasks to infer the posterior over task identity, as detailed in Sec. \ref{sec_exp_ablation} and pseudo-codes provided in Alg. \ref{alg:two_phase_distill}. Therefore, the effective number of transitions that are used as input into the task inference module during training are the same for our model and Contextual BCQ. 

Unless stated otherwise, for the remaining hyper-parameters, such as the maximum value of the perturbation, we use the default value in BCQ.

\subsection{Hyper-parameters of PEARL}

\begin{table}[H]
\small
\centering
    \caption{Hyper-parameters of PEARL}\label{table:hp-pearl}
    \begin{tabular}{ |l|l| }
            \hline
            {\bf Hyper-parameters} & {\bf Value} \\ 
            \hline
            Task inference module architecture & MLP with 3 hidden layers, 200 nodes each, ReLU activation \\
            
            \hline
    \end{tabular}
\end{table}

We use the default hyper-parameters as provided in the official implementation of PEARL. For completeness when discussing the hyper-parameters of our model, we provide the architecture of the task inference module in \autoref{table:hp-pearl}.

\subsection{Hyper-parameters of ablation studies of the full model}
\begin{table}[H]
\small
\centering
    \caption{Hyper-parameters of \textbf{No transition relabelling}}\label{table:hp-no-trans-relbel}
    \begin{tabular}{ |l|l| }
            \hline
            {\bf Hyper-parameters} & {\bf Value} \\ 
            \hline
            Number of sampled context sets $N$ & 10 \\
            \hline
            Context set size & 128 \\
            \hline
    \end{tabular}
\end{table}

\begin{table}[H]
\small
\centering
    \caption{Hyper-parameters of \textbf{No triplet loss}}\label{table:hp-no-triplet-loss}
    \begin{tabular}{ |l|l| }
            \hline
            {\bf Hyper-parameters} & {\bf Value} \\ 
            \hline
            Context set size & 64 \\
            \hline
    \end{tabular}
\end{table}

\begin{table}[H]
\small
\centering
    \caption{Hyper-parameters of \textbf{Neither}}\label{table:hp-neither}
    \begin{tabular}{ |l|l| }
            \hline
            {\bf Hyper-parameters} & {\bf Value} \\ 
            \hline
            Context set size & 128 \\
            \hline
    \end{tabular}
\end{table}

\autoref{table:hp-no-trans-relbel}, \autoref{table:hp-no-triplet-loss} and \autoref{table:hp-neither} provide the hyper-parameters for the ablated versions of our full model \textbf{No transition relabelling}, \textbf{No triplet loss}, and \textbf{Neither}, respectively. Without the transition relabelling techniques, \textbf{No transition relabelling} and \textbf{Neither} set the size of training context size to $128$ as Contextual BCQ to use the same effective number of transitions to infer the posterior over the task identity as our full model. Note that the remaining hyper-parameters of these methods are set to be the same as our full model, described in \autoref{table:hp-distillation}.

\subsection{Hyper-parameters when we initialize SAC with our multi-task policy}

\begin{table}[H]
\small
\centering
    \caption{Hyper-parameters of SAC when initialized by our multi-task policy}\label{table:hp-sac-init}
    \begin{tabular}{ |l|l| }
            \hline
            {\bf Hyper-parameters} & {\bf Value} \\ 
            \hline
            Q function architecture & MLP with 9 hidden layers, 1024 nodes each, ReLU activation \\
            Q function target smoothing rate & 0.005 \\
            policy target smoothing rate & 0.1 \\
            \hline
    \end{tabular}
\end{table}

The architecture of the Q function network is the same as the distilled Q function $Q_D$ in \autoref{table:hp-distillation}. The Q function target smoothing rate is the same as the standard SAC implementation \cite{haarnoja2018soft}. The policy target smoothing rate is searched over $\{0.005, 0.01, 0.1, 0.5\}$. For the SAC trained from random initialization baseline (Appendix \ref{sec_init_counterpart}), we also change the sizes of the value function to the same value in \autoref{table:hp-sac-init}. For the remaining hyper-parameters, we use the default hyper-parameter settings of SAC.

\clearpage

\section{Reward prediction ensemble}\label{sec_ensemble}
\begin{algorithm}[H]
	\caption{Training procedure of reward function approximator}\label{alg:reward_predict}
	{\bf Input}: data batch $\mathcal{B}_i$; $\hat{R}_{i,l}$ with randomly initialized parameters.
	\begin{algorithmic}[1]
        \For {a fixed number of iterations}
            \State Sample a transition $(s,a,r,s')$ from $\mathcal{B}_i$ 
            \State Obtain the predicted reward $\hat{r} = \hat{R}_{i,l}(s,a)$
            \State Update parameters of $\hat{R}_{i,l}$ to minimize $(\hat{r} - r)^2$ through gradient descent.
        \EndFor
	\end{algorithmic}
	{\bf Output}: trained reward function approximator $\hat{R}_{i,l}$
\end{algorithm}
\vspace{0.2 in}

\begin{algorithm}[H]
	\caption{Relabel transition from task $j$ to task $i$}\label{alg:relabel}
	{\bf Input}:an ensemble of learned reward functions $\{\hat{R}_{i,l}\}_l$; context set ${\bf c}_j = \left\{ \left(s_{j, t}, a_{j, t}, r_{j, t}, s_{j, t}^{\prime}\right) \right\}_t$ from task $j$, a threshold $\sigma_{\text{threshold}}$.
	\begin{algorithmic}[1]
        \State ${\bf c}_{j\rightarrow{i}} \leftarrow \{\}$
        \For {$t = 1, \ldots, |{\bf c}_j|$}
            \If {$\text{std}( \{\hat{R}_{i,l}(s_{j, t}, a_{j, t})\}_l) < \sigma_{\text{threshold}}$}
                \State 
                $\hat{R}_i(s_{j, t}, a_{j, t})\leftarrow\text{mean}( \{\hat{R}_{i,l}(s_{j, t},a_{j,t}\}_l)$
                \State Add $(s_{j, t}, a_{j, t}, \hat{R}_i(s_{j, t}, a_{j, t}), s_{j, t}^{\prime} )$ to ${\bf c}_{j\rightarrow{i}}$
            \EndIf
        \EndFor
	\end{algorithmic}
	{\bf Output}: relabelled transitions ${\bf c}_{j\rightarrow{i}}$
\end{algorithm}
\vspace{0.1 in}

In \autoref{sec_algo_triplet}, we propose to train a reward function approximator $\hat{R}_i$ for each training task $i$ to relabel the transitions from the other tasks. To increase the accuracy of the estimated reward, for each task $i$, we use an ensemble of learnt reward functions $\{\hat{R}_{i,l}\}_l$, where $i$ indexes the task and $l$ indexes the function in the ensemble. The training procedures for each reward function approximator in the ensemble are provided in Alg. \ref{alg:reward_predict}. 

The pseudo-code for generating relabelled context set ${\bf c}_{j\rightarrow{i}}$ from context set ${\bf c}_j$ of task $j$ is given in Alg. \ref{alg:relabel}. We use the output of the ensemble as an estimate of the epistemic uncertainty in the reward prediction \cite{chua2018deep}. Concretely, for each transition in ${\bf c}_j$, we only include it in the relabelled set ${\bf c}_{j\rightarrow{i}}$ if the standard deviations of the ensemble output is below a certain threshold (line 3). We also use the mean of the outputs as the estimated reward (line 4).

We conduct ablation study of the reward prediction ensemble in Appendix \ref{sec_ablate_ensemble}, where we show that the use of reward prediction ensemble improves the performance when initializing SAC with our multi-task policy. 

\clearpage

\section{Detailed pseudo-codes of the two-phases distillation procedures }\label{sec_detailed_distill}

In this section, we provide the detailed pseudo-code in Alg. \ref{alg:two_phase_distill} for the two-phases distillation procedures introduced in Sec. \ref{sec_algo}. The basic idea is that we first obtain single-task policy for each training task using BCQ. In the second phase, we distill the single-task policies into a multi-task policy by incorporating a task inference module. Note that the task inference module is trained by minimizing the Q value function distillation loss (Eq. \ref{eq_loss_qi}) and the triplet loss (Eq. \ref{eq_triplet_all_task}).

Line 1 describes the first phase of the two-phases distillation procedure. We use BCQ to learn a state-action value function $Q_i$, a candidate action generator $G_i$ and a perturbation generator $\xi_i$ for each training batch $\mathcal{B}_{i}$.

We next enter the second phase. We first sample context set ${\bf c}_i$ of size $N_{\text{context}}$ from $\mathcal{B}_{i}$, $i = 1, \ldots, K$ in line 3. Line 5-10 provide the procedures to calculate the triplet loss. For each task $i$, we relabel the reward of each transition in all the remaining context set ${\bf c}_j$ using $\hat{R}_i$ and obtain ${\bf c}_{j\rightarrow i}, \forall j\neq i$ in line 5. From the union of the relabelled context set $\cup_j {{\bf c}_{j \rightarrow{i}}}$, we sample a subset $\tilde{{\bf c}}_i$ of size $N_{\text{context}}$ in line 6. Denote transitions in $\tilde{{\bf c}}_i$ originated from ${\bf c}_j$ as ${\bf x}_{j\rightarrow{i}}$. Further denote transitions in ${\bf x}_{j\rightarrow{i}}$ before relabelling as ${\bf x}_j$, we thus have ${\bf x}_j \in {\bf c}_j$. These sets of transitions have the following relationships:
\begin{align}
    &   \cup_j {{\bf c}_{j \rightarrow{i}}} \xrightarrow{\text{Sample}} \tilde{{\bf c}}_i, \quad \tilde{{\bf c}}_i = \cup_j {{\bf x}_{j \rightarrow{i}}} \nonumber\\
    & {\bf x}_{j \rightarrow{i}} \in {\bf c}_{j \rightarrow{i}}, \quad  {\bf x}_j \xrightarrow{\text{Relabel}} {{\bf x}_{j \rightarrow{i}}}
\end{align}
To calculate the triplet loss for task $i$, in line 9 we sample a subset ${\bf c}_{i,j}$ with the same number of transitions as ${\bf x}_j$ from ${\bf c}_i$, i.e. $|{\bf c}_{i, j}| = |{\bf x}_j|$ for each $j\neq i$. Therefore, the triplet loss for task $i$ can be given by Eq. \ref{eq_real_triplet_loss_i}.

Line 11-13 provide the procedures to infer the task identity for each task $i$. We use the union of the context set ${\bf c}_i$ and the relabeled context set $\tilde{{\bf c}}_i$ sampled from $\cup_j {{\bf c}_{j \rightarrow{i}}}$ to infer the posterior $q_{\phi}({\bf z}|\{{\bf c}_i, \tilde{{\bf c}}_i\})$ over task identity. We next sample the task identity ${\bf z}_i$ from $q_{\phi}({\bf z}|\{{\bf c}_i, \tilde{{\bf c}}_i\})$. 

To calculate the distillation loss of each distilled function, in line 14 we sample the training batch of $N$ transitions from $\mathcal{B}_{i}$. With ${\bf z}_i$ and the training transition batch, we can calculate the value function distillation loss $\mathcal{L}^i_{Q}$ of task $i$ using Eq. \ref{eq_real_qfunc_loss_i}. To calculate the distillation loss of the candidate action generator $G_D$ and perturbation generator $\xi_D$ of task $i$, we first sample $N$ noises $\nu_t$ from the standard Gaussian distribution $\mathcal{N}(0, 1)$ in line 16. In line 17, we then obtain the candidate actions $\hat{a}_t = G_i(s_t, \nu_t)$ for each state $s_t$ in the training batch. The calculations to derive $\mathcal{L}^i_{G}$ and $\mathcal{L}^i_{\xi}$ for task $i$ follow Eq. \ref{eq_real_g_loss_i} and Eq. \ref{eq_real_xi_loss_i}, respectively.

After repeating the procedures for all the training tasks, in line 21-24 we average the losses across tasks and obtain $\mathcal{L}_{triplet}$, $\mathcal{L}_{Q}$, $\mathcal{L}_{G}$, and $\mathcal{L}_{\xi}$. At the end of each iteration, we update $\theta$ and $\phi$ by minimizing $\mathcal{L} = \mathcal{L}_{triplet} + \mathcal{L}_Q + \mathcal{L}_G + \mathcal{L}_\xi$ in line 25.

\begin{algorithm}[t]
    \caption{Two-phases distillation procedure with novel triplet loss design}\label{alg:two_phase_distill}
    {\bf Input}: Batches $\{\mathcal{B}_{i}\}_{i=1}^K$; trained reward function $\{\hat{R}_i\}_{i=1}^K$; randomly initialized $Q_D$, $G_D$ and $\xi_D$ jointly parameterized by $\theta$; task inference module $q_\phi$ with randomly initialized $\phi$; context set size $N_{\text{context}}$; training batch size $N$; triplet margin $m$
    \begin{algorithmic}[1]
    
        \State Learn single task policy $Q_i$, $G_i$, and $\xi_i$ from each data batch $\mathcal{B}_{i}$ using BCQ, $\forall i$
        \Repeat
            \State Sample context set ${\bf c}_i$ from $\mathcal{B}_{i}, \forall i$
             \For {$i = 1, \ldots, K$}
                \State Obtain the relabelled context set ${\bf c}_{j\rightarrow i}$ from ${\bf c}_j$ with $\hat{R}_i$ according to Alg. \ref{alg:relabel}, $\forall j\neq i$
                \State Sample a subset of relabelled context set $\tilde{{\bf c}}_i$: $\tilde{{\bf c}}_i \sim \cup_j {{\bf c}_{j \rightarrow{i} } }$,  $|\tilde{{\bf c}}_i| = N_{\text{context}}$
                \State Denote transitions in $\tilde{{\bf c}}_i$ originated from ${\bf c}_j$ as ${\bf x}_{j\rightarrow{i}}$
                \State Denote transitions in ${\bf x}_{j\rightarrow{i}}$ before relabelling as ${\bf x}_j$, ${\bf x}_j \in {\bf c}_j$
                \State Sample a subset ${\bf c}_{i,j}$ from ${\bf c}_i$ with $|{\bf c}_i| = |{\bf x}_j|$, $\forall j\neq i$ 
                \State Calculate the triplet loss $\mathcal{L}^i_{triplet}$ 
                \begin{equation}\label{eq_real_triplet_loss_i}
                    \mathcal{L}^i_{triplet} = \frac{1}{K-1} \sum_{j=1, j \neq i}^{K} \left[d(q_{\phi}\left({\bf x}_{j\rightarrow{i}}), q_{\phi}\left({\bf c}_{i,j}\right)\right)
                    - d(q_{\phi}\left({\bf x}_{j\rightarrow{i}}\right), q_{\phi}\left({\bf x}_{j}\right)) + m\right]_{+}
                \end{equation}
                \State Combine ${\bf c}_i$ and $\tilde{{\bf c}}_i$ to form the new context set $\{{\bf c}_i, \tilde{{\bf c}}_i\}$ 
                \State Infer the posterior $q_{\phi}({\bf z}|\{{\bf c}_i, \tilde{{\bf c}}_i\})$ over task identity from $\{{\bf c}_i, \tilde{{\bf c}}_i\}$
                \State Sample task identity ${\bf z}_i \sim q_{\phi}({\bf z}|\{{\bf c}_i, \tilde{{\bf c}}_i\})$
                \State Sample training batch: $\{(s_{t}, a_{t}, r_{t}, s'_{t})\}^N_{t=1}$
                \State Calculate the value function distillation loss
                \begin{equation}\label{eq_real_qfunc_loss_i}
                \mathcal{L}^i_Q =  \frac{1}{N}\sum^N_{t=1}\left[Q_i(s_t, a_t) - Q_D(s_t, a_t, {\bf z}_i) )^2\right] + \beta\text{KL}(q_{\phi}({\bf z}|\{{\bf c}_i, \tilde{{\bf c}}_i\}) ||  \mathcal{N}(0,1))
                \end{equation}
                \State Sample $N$ noises: $\nu_t \sim \mathcal{N}(0,1)$, $t = 1,\ldots, N$
                \State Obtain candidate action from $G_i$: $\hat{a}_t = G_i(s_t, \nu_t)$, $t = 1,\ldots, N$
                \State Calculate the candidate action generator distillation loss
                \begin{equation}\label{eq_real_g_loss_i}
                    \mathcal{L}^i_G = \frac{1}{N}\sum^N_{t=1}\left[||\hat{a}_t - G_D(s_t, \nu_t, \bar{\bf z}_i)||^2\right]
                \end{equation}
                \State Calculate the perturbation generator distillation loss
                \begin{equation}\label{eq_real_xi_loss_i}
                    \mathcal{L}^i_\xi = \frac{1}{N}\sum^N_{t=1}\left[||\xi_i(s_t, \hat{a}_t) - \xi_D(s_t, \hat{a}_t, \bar{\bf z}_i) || ^ 2\right]
                \end{equation}
            \EndFor
        
        \State Calculate $\mathcal{L}_{triplet} = \frac{1}{K}\sum^K_{t=1}\mathcal{L}^i_{triplet}$
        \State Calculate $\mathcal{L}_Q = \frac{1}{K}\sum^K_{t=1}\mathcal{L}^i_Q$
        \State Calculate $\mathcal{L}_G = \frac{1}{K}\sum^K_{t=1}\mathcal{L}^i_G$
        \State Calculate $\mathcal{L}_\xi = \frac{1}{K}\sum^K_{t=1}\mathcal{L}^i_\xi$
        \State Update $\theta, \phi$ to minimize $\mathcal{L} = \mathcal{L}_{triplet} + \mathcal{L}_Q + \mathcal{L}_G + \mathcal{L}_\xi$
        \Until{Done}
    \end{algorithmic}
\end{algorithm}

\clearpage

\section{Action selection and evaluation of the multi-task policy }\label{sec_action}

\begin{figure}[H]
\centering
\includegraphics[width=0.7\textwidth]{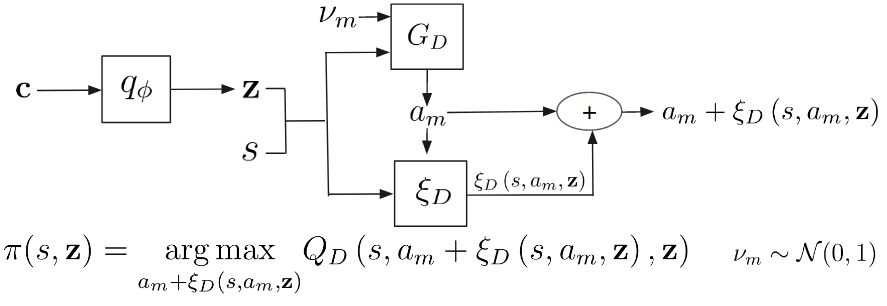}
\caption{Action selection. Given context set ${\bf c}$, $q_\phi$ infer the posterior over task identity, from which we sample the task identity ${\bf z}$. With the task identity ${\bf z}$, $G_D$ generates multiple candidate actions $a_m$ for state $s$. $\xi_D$ generates small corrections $\xi_D\left(s, a_{m}, {\bf z}\right)$ for the candidate actions $a_m$. The policy takes the corrected action $a_m + \xi_D\left(s, a_{m}, {\bf z}\right)$ with the highest value as estimated by $Q_D$.}

\label{fig:action-appendix}
\end{figure}

In this section, we will describe the action selection procedures from the multi-task policy as shown in Fig. \ref{fig:action-appendix}, and how we evaluate its performance.

Sampling action given a state from the multi-task policy is similar to the procedures of BCQ (\autoref{sec_action_bcq}). The main difference is that the networks also take an inferred task identity ${\bf z}$ as input. Concretely, given a state $s$, the distilled candidate action generator $G_D$ generates multiple candidate actions $\{a_m = G_D(s, \nu_m, {\bf z})\}_m$ with random noise $\nu_m\sim \mathcal{N}(0, 1)$. The distilled perturbation generator $\xi_D$ generates a small correction term $\xi_D(s, a_m, {\bf z})$ for each state-candidate action pair. We take the corrected action with the highest value as estimated by the distilled value function $Q_D$. The action selection procedures can be summarized by:
\begin{align}\label{eq_action_selection}
   \pi(s, {\bf z})=\underset{a_{m}+\xi_D\left(s, a_{m}, {\bf z}\right)}{\arg \max } Q_D\left(s, a_{m}+\xi_D\left(s, a_{m}, {\bf z}\right), {\bf z}\right), \, \left\{a_{m}=G_D\left(s, \nu_{m}, {\bf z}\right)\right\}_{m}, \, \nu_{m} \sim \mathcal{N}(0,1).
\end{align}

We elaborate the evaluation procedures in Alg. \ref{alg:test}. When testing on a new task, we do not have the ground truth task identity or any transition from the task to infer the task identity. We thus sample the initial task identity from the standard Gaussian prior in line 1. The task identity is kept fixed for the duration of the first episode. Afterwards, we use the collected transitions to infer the posterior and sample new task identities before each new episode as described in line 3. When calculating the average episode returns, we do not count the first two episodes’ returns as what is done in \cite{rakelly2019efficient}. 
\vspace{0.1 in}
\begin{algorithm}[t]
	\caption{Evaluation procedures of our model}\label{alg:test}
	{\bf Input}: unseen task $\mathcal{M}$; learned multi-task policy
	\begin{algorithmic}[1]
        \State Initialize context set ${\bf c} \leftarrow \{\}$; initialize $q_\phi({\bf z}|{\bf c}) = \mathcal{N}(0,1)$
        \Repeat
            \State Sample task identity ${\bf z} \sim q_\phi({\bf z}|{\bf c})$.
            \State Collect one episode of transitions $\{(s_t, a_t, r_t, s'_t)\}_t$ from task $\mathcal{M}$ with multi-task policy conditioned on ${\bf z}$.
            \State Add $\{(s_t, a_t, r_t, s'_t)\}_t$ to {\bf c}.
        \Until{Done}
	\end{algorithmic}
	{\bf Output}: average episode returns, not counting the first two episodes
\end{algorithm}

\clearpage

\section{On Modifying the original HumanoidDir task distribution}
\label{sec_humanoid}

We are concerned the original HumanoidDir task distribution is not suitable as a benchmark for multi-task RL because a policy trained from a single task can already obtain the optimal performance on unseen tasks. In particular, we train BCQ with transitions from one task and it obtains a similar return, as measured on unseen tasks ($993 \pm 33$), to SAC trained from scratch separately for each task ($988 \pm 19$).

In the HumanoidDir task distribution, each task is defined by a target running direction. The intended task is for the agent to run with maximal velocity in the target direction. The reward of each task can be defined as below:
\begin{align}
    R(s,a,s') = \text{alive}\_\text{bonus} & + \alpha * \text{achieved}\_\text{velocity} \cdot \text{target}\_\text{direction}\nonumber \\ & - \text{quad}\_\text{ctrl}\_\text{cost} - \text{quad}\_\text{impact}\_\text{cost},
\end{align}
where $\cdot$ denotes the inner product. Note that the two cost terms tend to be very small thus it will be reasonable to omit them in analysis. The $\text{alive}\_\text{bonus}$ is the same across different tasks and is a constant. The $\text{target}\_\text{direction}$ is different across tasks. $\alpha$ weights their relative contribution to the reward. If $\alpha$ is too small, the reward is dominated by the constant $\text{alive}\_\text{bonus}$. In this case, to achieve good performance, the agent does not need to perform the intended task. In other word, the agent does not need to infer the task identity to obtain good performance and only needs to remain close to the initial state while avoiding terminal states to maximize the episode length.

Prior works that use HumanoidDir set $\text{alive}\_\text{bonus} = 5.0$ and $\alpha = 0.25$. With such a small value for the reward coefficient $\alpha$, the reward is dominated by the $\text{alive}\_\text{bonus}$. We provide video to illustrate that in different tasks, the SAC-trained single-task policies display similar behaviors even though the different tasks have different running directions\footnote{Videos are provided: \url{https://www.youtube.com/channel/UCWrYNNRgZzqxnhfOYbgNmkA}}. In most tasks, the SAC-trained policy controls the Humanoid to stay upright near the initial state, which is enough to obtain high performance. If a single policy that controls the agent to stay upright can achieve high performance in all tasks sampled from this task distribution, we argue that the learned multi-task policy in this task distribution can achieve near-optimal performance across tasks without the need to perform accurate task inference. In other word, this task distribution is not suitable to demonstrate the test-time task inference challenge identified in our work.


Therefore, we set $\alpha = 1.25$, which is the value used in the OpenAI implementation of Humanoid\footnote{OpenAI implementation of Humanoid-v2 is provided here \url{https://github.com/openai/gym/blob/master/gym/envs/mujoco/humanoid.py}}, and denote the modified task distribution as HumanoidDir-M. As is shown in the video, the SAC-trained agent in our case runs with significant velocity in the target direction. The optimal behaviors among the different tasks are thus sufficiently different such that the multi-task policy needs to infer the task identity to obtain high performance.

\clearpage

\section{Details of the environmental settings and baseline algorithms}\label{sec_env_baseline}

In this section, we will first provide the details of environmental settings in Appendix \ref{sec_env_detail}, and then describe the baseline algorithms we compare against in Sec \ref{sec_exp}. We explain PEARL in Appendix \ref{sec_pearl} and Contextual BCQ in Appendix \ref{sec_contextual_bcq}.

\subsection{Environment setups}\label{sec_env_detail}


We construct the task distribution UmazeGoal-M by modifying the maze-umaze-dense-v1 from D4RL. We always reset the agent from the medium of the U shape maze, while the goal locations is randomly initialized around the two corners of the maze.

The episode length is 1000 for HalfCheetahVel, which is the episode length commonly used when model-free algorithms are tested in the single-task variant of these task distributions. We use the same episode length 300 as D4RL for UmazeGoal-M. In the remaining task distributions, we set the episode length to be 200 due to constrained computational budget.

\autoref{table:sac_bcq_perform} provides details on each task distribution, including the number of training tasks and number of testing tasks. Note that the set of training tasks and the set of testing tasks do not overlap. The column "Interactions" specifies the number of transitions available for each task.
With the selected number of interactions with the environment, we expect the final performance of training SAC in each task to be slightly below the optimal performance. In other word, we do not expect the batch data to contain a large amount of trajectories with high episode returns.

\begin{table}[H]
\vspace{0.2 in}
\centering
\small
\begin{tabular}{lccccccc}
\hline 
& Num train tasks & Num test tasks & Interactions & SAC returns & BCQ returns\\
\hline 
HalfCheetahVel  & 10 & 8 & 60K & $-121.3_{\pm 35.3}$ & $-142.7_{\pm 29.9}$ \\ 

AntDir  & 10 & 8 & 200K & $920.9_{\pm 85.4}$ & $956.6_{\pm 83.8}$ \\ 

AntGoal & 10 & 8 & 300K & $-99.6_{\pm 33.9}$ & $-127.8_{\pm 36.7}$ \\ 

WalkerParam  & 30 & 8 & 300K & $671.1_{\pm 106.4}$ & $692.6_{\pm 97.0}$ \\

HumanoidDir-M  & 10 & 8 & 600K & $2116.1_{\pm 388.6}$ & $2190.9_{\pm 370.9}$ \\

UmazeGoal-M & 10 & 8 & 30K & $252.8_{\pm 6.5}$ & $258.3_{\pm 9.1}$ \\ 

\hline
\end{tabular}
\vspace{0.1 in}
\caption{Details of the experimental settings} 
\label{table:sac_bcq_perform}
\end{table}
 
\subsection{PEARL under Batch RL setting}\label{sec_pearl}

Our works are very much inspired by PEARL \cite{rakelly2019efficient}, which is the state-of-the-art algorithm designed for optimizing the multi-task objective in various MuJoCo benchmarks. By including the results for PEARL, we demonstrate that conventional algorithms that require interaction with the environment during training does not perform well in the Multi-task Batch RL setting, which motivates our work.

To help readers understand the changes we made to adapt PEARL to the Batch RL setting, we reuse the notations from the original PEARL paper in this section. Detailed training procedures are provided in Algorithm \ref{alg:pearl}. Without the privilege to interact with the environment, PEARL proceeds to sample the context set ${\bf c}^i$ from the task batch $\mathcal{B}_i$ in line 5. The task inference module $q_\phi$, parameterized by $\phi$ takes as input the context set ${\bf c}^i$ to infer the posterior $q_\phi({\bf z}|{\bf c}^i)$. In line 6, we sample the task identity ${\bf z}_i$ from $q_\phi({\bf z}|{\bf c}^i)$. In line 7-9, the task identity ${\bf z}_i$ combined with the RL mini-batch $b^i$ is further input into the SAC module. For task $i$, $\mathcal{L}^i_{actor}$ defines the actor loss, and $\mathcal{L}^i_{critic}$ defines the critic loss. $\mathcal{L}^i_{KL}$ constrains the inferred posterior $q({\bf z} | {\bf c}^i)$ over task identity from context set ${\bf c}^i$ to stay close to the prior $r({\bf z})$. As shown in line 11, gradients from minimizing both $\mathcal{L}^i_{critic}$ and $\mathcal{L}^i_{KL}$ are used to train the task inference module $q_\phi$. We refer the readers to the PEARL paper for detailed definitions of these loss functions. 

In PEARL, the context set is sampled from a replay buffer of recently collected data, while the training RL mini-batches (referred to as the RL batches in PEARL) is sampled uniformly from the replay buffer. This is not possible in the Multi-task Batch RL setting since all transitions are collected prior to training and are ordered arbitrarily. There is thus not a well-defined notion of "recently collected data". 

\begin{algorithm}[t]
\begin{algorithmic}[1]
\State \textbf{Require:} Batches $\{\mathcal{B}_i\}^K_{i=1}$, learning rates $\alpha_1, \alpha_2, \alpha_3$ 
\While{not done}
    \For{step in training steps}
    \For{$i = 1,\ldots,K$}
    \State Sample context ${\bf c}^i \sim \mathcal{B}^i$ and RL batch $b^i \sim \mathcal{B}^{i}$
    \State Sample ${\bf z} \sim q_{\phi}({\bf z} | {\bf c}^i)$
    \State $\mathcal{L}^i_{actor} = \mathcal{L}_{actor}(b^i, {\bf z})$
    \State $\mathcal{L}^i_{critic} = \mathcal{L}_{critic}(b^i, {\bf z})$
    \State $\mathcal{L}^i_{KL} = \beta D_{\text{KL}}(q({\bf z} | {\bf c}^i) || r({\bf z}))$
    \EndFor
    \State $\phi \gets \phi - \alpha_1 \nabla_\phi \sum_i \left(\mathcal{L}^i_{critic} + \mathcal{L}^i_{KL}\right)$
    \State $\theta_{\pi} \gets \theta_{\pi} - \alpha_2  \nabla_\theta \sum_i \mathcal{L}^i_{actor}$
    \State $\theta_{Q} \gets \theta_{Q} - \alpha_3  \nabla_\theta \sum_i \mathcal{L}^i_{critic}$
    \EndFor
\EndWhile
\end{algorithmic}
\caption{{ PEARL under Multi-task Batch RL setting (modified from Algorithm 1 in PEARL)}}
\label{alg:pearl}
\end{algorithm}

\subsection{Contextual BCQ}\label{sec_contextual_bcq}

\begin{algorithm}[t]
  \caption{Contextual BCQ (modified from Algorithm 1 in BCQ \cite{fujimoto2019off})}
  \label{algorithm:contextual_BCQ}
\begin{algorithmic}[1]
	\State \textbf{Input:} Batches $\{\mathcal{B}_m\}^K_{m=1}$, horizon $T$, target network update rate $\tau$, mini-batch size $N$, max perturbation $\Phi$, number of sampled actions $n$, minimum weighting $\lambda$. 
	\State Initialize task inference module $q_\psi$ Q-networks $Q_{\theta_1}, Q_{\theta_2}$, perturbation network $\xi_\phi$, and VAE $G_\w = \{E_{\w_1}, D_{\w_2}\}$, with random parameters $\psi$, $\theta_1$, $\theta_2$, $\phi$, $\omega$, and target networks $Q_{\theta'_1}, Q_{\theta'_2}$, $\xi_{\phi'}$ with $\theta'_1 \leftarrow \theta_1, \theta'_2 \leftarrow \theta_2$, $\phi' \leftarrow \phi$.
	\Repeat
	\For{$m = 1,\ldots, K$}
    	\State Sample $N$ transitions $\left\{\left(s, a, r, s^{\prime}\right)_{t} \right\}_t$ from each $\mathcal{B}_m$
    	\State Sample context set ${\bf c}_m$ from $\mathcal{B}_m$
    	\State Sample task identity ${\bf z}_m$ from the inferred posterior $q_\psi({\bf z}|{\bf c}_m)$
    	\State $\mu, \s = E_{\omega_1}({s}, a, {\bf z}_m), \quad \tilde a = D_{\omega_2}({s}, \nu, {\bf z}_m), \quad \nu \sim \N(\mu, \sigma)$ 
    	\State $\omega \leftarrow \argmin_\omega \sum (a - \tilde a)^2 + D_{\text{KL}}(\N(\mu, \s)||\N(0,1))$ 
    	\State Sample $n$ actions: $\{a_i \sim G_\omega(s', {\bf z}_m)\}_{i=1}^n$ 
    	\State Set value target $y$ (Eq. \ref{eq_qtraget})
    	\State $\theta \leftarrow \argmin_\theta \sum (y - Q_\theta({s},a, {\bf z}_m))^2$ 
    	\State $\phi \leftarrow \argmax_\phi \sum Q_{\theta_1}({s},a + \xi_\phi({s}, a, {\bf z}_m, \Phi), {\bf z}_m), a \sim G_\omega({s}, {\bf z}_m)$ 
    	\State $\psi \leftarrow \argmin_\psi \sum (y - Q_\theta({s},a, {\bf z}_m))^2$
    	\State Update target networks: $\theta'_i \leftarrow \tau \theta + (1 - \tau) \theta'_i$
    	\State $\phi' \leftarrow \tau \phi + (1 - \tau) \phi'$
	\EndFor
	\Until{iterates for $T$ times}
\end{algorithmic}
\end{algorithm}

We reuse the notations from the original BCQ paper \cite{fujimoto2019off} to help reader understand how we modify modify BCQ to train multi-task policy by incorporating a task inference module. We refer to this method as Contextual BCQ and use it to serve as our baseline methods. By comparing this baseline, we argue that the problem we are facing cannot be solved by simply combining the current Batch RL algorithm with a simple task inference module. Next, we will start by providing a brief introduction of the training procedures of BCQ.

Batch Constrained Q-Learning (BCQ) is a Batch RL algorithm that learns the policy from a fixed data batch without further interaction with the environment \cite{fujimoto2019off}. By identifying the extrapolation error, BCQ restricts the action selection to be close actions taken in batch. Specifically, it trains a conditional variational auto-encoder $G$ \cite{kingma2013auto} to generate candidate actions that stay close to the batch for each state $s$. A perturbation model $\xi$ will generate a small additional correction term to induce limited exploration for each candidate action in the range $[-\Phi, \Phi]$. The perturbed action with the highest state action value as estimated by a learned value function $Q$ will be selected. 

By modifying BCQ to incorporate incorporate module detail, the training procedures of Contexual BCQ can be detailed in Alg. \ref{algorithm:contextual_BCQ}. As the original BCQ algorithm, we maintain two separate Q function networks $Q_{\theta_1}$, $Q_{\theta_2}$ parameterized by $\theta_1, \theta_2$, a generative model $G_\omega = \{E_{\omega_1}, D_{\omega_2}\}$ parameterized by $\omega$, where $E_{\omega_1}, D_{\omega_2}$ are the encoder and decoder, and a perturbation generator $\xi_\phi$ parameterized by $\phi$. For the $Q_{\theta_1}$, $Q_{\theta_2}$ and $\xi_\phi$, we also maintain their corresponding target networks $Q_{\theta'_1}, Q_{\theta'_2}$ and $\xi_{\phi'}$. Compared with the original BCQ, all these networks will take in the inferred task identity ${\bf z}$ as an extra input, which is generated by the task inference module as $q_\psi$ parameterized by $\psi$.

We use $m$ to index the task. From each task batch $\mathcal{B}_m$, in line 5, we sample a context set ${\bf c}_m$ and $N$ transitions $\{s_{t}, a_{t}, r_{t}, s_{t}^{\prime})\}_t$, where $t$ indexes the transition. For simplicity, we denote the transitions with the shorthanded $\{(s, a, r, s^{\prime})_{t}\}_t$. In line 7, $q_\psi$ takes as input the context set ${\bf c}_m$ and infer a posterior $q_\psi({\bf z}|{\bf c}_m)$ over the task identity, from which we sample a task identity ${\bf z}_m$.

Line 8-9 provide the procedures to train the generative model $G_\omega = \{E_{\omega_1}, D_{\omega_2}\}$. Specifically, $E_{\omega_1}$ takes as the input the state-action pair $(s,a)$ and task identity ${\bf z}_m$ and output the mean $\mu$ and variance $\sigma$ of a Gaussian distribution $\mathcal{N}(\mu, \sigma)$. That is, $\mu, \sigma = E_{\omega_1}(s, a, {\bf z}_m)$. From $\mathcal{N}(\mu, \sigma)$, we sample a noise $\nu$ and input it to the decoder $D_{\omega_2}$ together with $s$ and ${\bf z}_m$ to obtain the reconstructed action $\tilde a = D_{\omega_2}(s, \nu, {\bf z}_m)$. We train $G_\omega$ by minimizing 
\begin{equation}
    \sum (a - \tilde a)^2 + D_{\text{KL}}(\N(\mu, \s)||\N(0,1)).
\end{equation}
Line 10-12 provide the procedures to train the Q value functions. For each next state $s'$ in the training batch, we can obtain $n$ candidate actions $\{a_i \sim G_\omega(s', {\bf z}_m)\}_{i=1}^n$ from the generative model $G_\omega$. This is done by sampling $n$ noises from the prior $\N(0, 1)$ and input to decoder $D_{\omega_2}$ together with $s'$, as shown in line 10. For each of the candidate action $a_i$, the perturbation model $\xi_\phi$ will generate a small correction term $\xi_\phi(s', a_i, {\bf z}_m, \Phi)\in[-\Phi, \Phi]$. We denote the perturbed actions as $\{a_i = a_i + \xi_\phi(\hat{s}',a_i, {\bf z}_m, \Phi)\}_{i=1}^n$. Therefore, the learning target for both of the Q function network is given by 
\begin{equation}\label{eq_qtraget}
    y = r+\gamma \max_{a_i}\left[\lambda \min _{j=1,2} Q_{\theta_{j}^{\prime}}\left(s^{\prime}, a_{i}, {\bf z}_m\right)+(1-\lambda) \max _{j=1,2} Q_{\theta_{j}^{\prime}}\left(s^{\prime}, a_{i}, {\bf z}_m\right)\right],
\end{equation}
where $a_i$ is selected from the set of perturbed actions and the minimum weighting $\lambda$ can be set to control the overestimation bias and future uncertainty. We also use Eq. \ref{eq_qtraget} to train the task inference module $q_\psi$ in line 14. 

In line 13,  $\xi_\phi$ is trained to generate a small perturbation term in range $[-\Phi, \Phi]$ so that the perturbed candidate actions $a + \xi_\phi(s, a, {\bf z}_m, \Phi)$ can maximize the state action value estimated by the Q function. Note that the candidate actions $a$ are output by the generative model $ G_\omega$. The loss function to train $\xi_\phi$ thus can be formulated as
\begin{equation}
    \sum Q_{\theta_1}(s,a + \xi_\phi(s, a, {\bf z}_m, \Phi), {\bf z}_m),\quad a \sim G_\omega(s, {\bf z}_m)
\end{equation}

\clearpage

\section{Additional experimental results}\label{sec_ablation}
In addition to the results already presented in Sec. \ref{sec_exp_ablation}, this section presents more experimental results to further understand different design choices of our model. We evaluate the performance of our model when we set different values of the \textit{triplet margin} in Appendix \ref{sec_margin}. We present the ablation study of the reward ensemble in Appendix \ref{sec_ablate_ensemble}.

\subsection{Ablation study of the triplet margin}\label{sec_margin}

Recall the triplet loss for task $i$ defined in Sec. \ref{sec_algo_triplet}, 
\begin{equation*}
    \mathcal{L}^i_{triplet} = \frac{1}{K-1} \sum_{j=1, j \neq i }^{K} \bigg[\explain{d \big(q_{\phi}\left({\bf c}_{j\rightarrow{i}}\big), q_{\phi}\left({\bf c}_{i}\right)\right) \quad}{Ensure ${\bf c}_{j\rightarrow{i}}$ and ${\bf c}_{i}$ infer \textit{similar} task identities \quad}
    - \explain{ \quad d\big(q_{\phi}\left({\bf c}_{j\rightarrow{i}}\right), q_{\phi}\left({\bf c}_{j}\right)\big) \quad}{Ensure ${\bf c}_{j\rightarrow{i}}$ and ${\bf c}_{j}$ infer \textit{different} task identities} + \quad m\bigg]_{+},
\end{equation*}
where ${\bf c}_{j\rightarrow i}$ denotes the set of transitions relabelled by the reward function $\hat{R}_i$ of task $i$, and ${\bf c}_{i}$ denotes the context set for task $i$. We include a positive term $m$ referred to as \textit{triplet margin} when calculating the triplet loss. With this term, we expect that $d \big(q_{\phi}\left({\bf c}_{j\rightarrow{i}}\big), q_{\phi}\left({\bf c}_{i}\right)\right)$ is at least smaller than  $d \big(q_{\phi}\left({\bf c}_{j\rightarrow{i}}\big), q_{\phi}\left({\bf c}_{j}\right)\right)$ by $m$.

\begin{figure*}[t]
\centering
  \makebox[\textwidth]{
  
  \begin{subfigure}{0.30\paperwidth}
    \includegraphics[width=\linewidth]{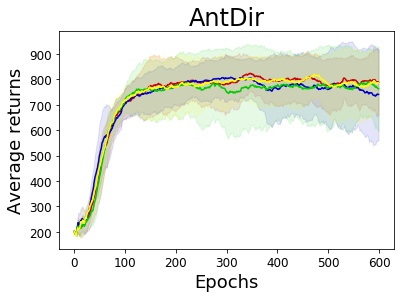}
  \end{subfigure}
  
  \begin{subfigure}{0.30\paperwidth}
    \includegraphics[width=\linewidth]{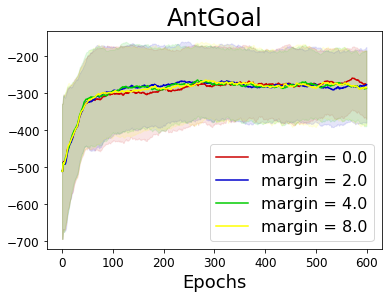}
  \end{subfigure}
}
  
  \makebox[\textwidth]{
  \begin{subfigure}{0.30\paperwidth}
    \includegraphics[width=\linewidth]{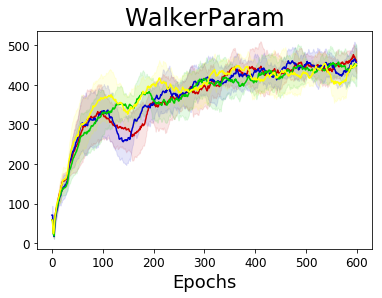}
  \end{subfigure}
  
  \begin{subfigure}{0.30\paperwidth}
    \includegraphics[width=\linewidth]{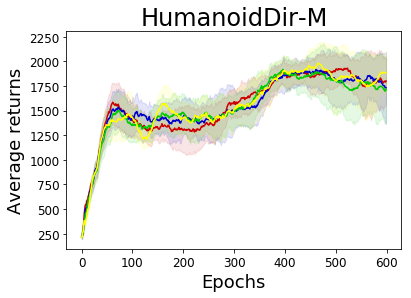}
  \end{subfigure}
  }
    
  \caption{Ablation study of the triplet margin on four task distributions. 
  The horizontal axis indicates the number of epochs. 
  The vertical axis indicates the average episode return.
  The shaded areas denote one standard deviation. }
\label{fig:triplet_margin}
\end{figure*}

Here, we examine how the performance of our algorithm changes when varying the value of the triplet margin $m$. Specifically, we set $m = 0.0, 2.0, 4.0, 8.0$ and show the results on the five task distributions. As can be seen in \autoref{fig:triplet_margin}, the performance of our model is in-sensitive to the value of the triplet loss margin. 

\subsection{Ablation study of reward prediction ensemble}\label{sec_ablate_ensemble}

\begin{figure}[t]
    \centering
    \includegraphics[width=0.5\textwidth]{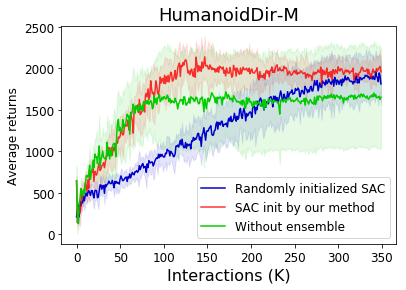}
    \caption{Results on HumanoidDir-M. When we use the multi-task policy trained without a reward ensemble to initialize SAC, performance has higher variance and converges to a lower value compared to using the ensemble. The horizontal axis indicates the number of training epochs. The vertical axis indicates the average episode return. The shaded areas denote one std.}
    \label{fig:without_ensemble}
\end{figure}

In Sec. \ref{sec_ensemble}, we describe the use of a reward ensemble to increase reward prediction accuracy when relabelling transitions. In this section, we demonstrate the benefit of the reward ensemble. Recall that in Sec. \ref{sec_exp_init_sac}, we use the trained multi-task policy as an initialization when further training is allowed on the unseen tasks. Using the reward ensemble when training the multi-task policy leads to higher performance when the trained multi-task policy is used as an initialization. Training our model without using a reward ensemble means we use one instead of an ensemble of networks to approximate the reward function for each training task. As shown in \autoref{fig:without_ensemble}, if the multi-task policy is trained without using the reward ensemble, when the multi-task policy is used as an initialization, the performance has high variance and has smaller asymptotic value.

\clearpage

\section{Details of using multi-task policy to initialize SAC}\label{sec_details_sac_init}

In this section, we provide more detailed illustrations of using the learned multi-task policy to initialize training on unseen tasks (Sec. \ref{sec_exp_init_sac}). We provide the pseudo-code for the SAC initialized by our methods in Appendix \ref{sec_sac_our_methods}. To demonstrate that the acceleration of convergence is really thanks to the transferability of our multi-task policy, we compare it against the variation of SAC, where we train the identically initialized two Q functions with different different mini-batch sampled from the replay buffer and also maintain a target policy network to stabilize the training process.

\subsection{Pseudo-code for SAC initialized by our methods}\label{sec_sac_our_methods}

To help readers understand the changes we made, we reuse the notations from the original SAC paper \cite{haarnoja2018soft} in this section. We first provide the pseudo-codes for initializing the new single-task policy $\pi_{\psi}$ parameterized by $\psi$ via imitation learning procedures in Alg. \ref{alg:imitate}. The whole training procedures are detailed in Alg. \ref{alg:sac_our_method} by modifying pseudo-code provided in \cite{SpinningUp2018}.

As is commonly done, the policy $\pi_{\psi}$ outputs the mean $\mu_s$ and $\Sigma_s$ of a Gaussian distribution for each state $s$, which characterizes the pdf of action selection, i.e. $p(a|s)\sim \mathcal{N}(\mu_s, \Sigma_s))$. In line 1 of Alg. \ref{alg:imitate}, we first collect 10K transitions using the multi-task policy in the unseen testing task $\mathcal{M}$. In line 2-7, We train $\pi_{\psi}$ to maximize the log likelihood of the action selections inside the collected data. Note that we also infer a task identity ${\bf z}$ from the 10K transitions in line 1. Its usage will be illustrated next. 

With the initialized single-task policy $\pi_\psi$ and inferred the task identity ${\bf z}$, we now turn to elaborate the whole process of the SAC initialized by the learnt multi-task policy. We detail the training procedures in Alg. \ref{alg:sac_our_method}. Compared with the standard SAC, in line 1 we initialize both the two Q function networks $Q_{\theta_1}$ and $Q_{\theta_2}$ identically with $Q_D$. However, in line 11-15, we train them using different batch data $B_1$ and $B_2$ sampled from the replay buffer $\mathcal{D}$ to stabilize the training process \cite{van2016deep}. In addition to maintain target networks $Q_{\theta_{\text{targ},1}}$ and $Q_{\theta_{\text{targ},2}}$ for each Q function, we also maintain a target policy network $\pi_{\psi_{\text{targ}}}$.

\begin{algorithm}[t]
	\caption{Imitation procedures}\label{alg:imitate}
	{\bf Input}: Unseen testing task $\mathcal{M}$; trained multi-task policy; randomly initialized single-task policy $\pi_\psi$;
	\begin{algorithmic}[1]
        \State Sample 10K transitions $\mathcal{R} = \{(s_t, a_t, r_t, s_t')\}_t$ from $\mathcal{M}$ using the multi-task policy and infer the task identity ${\bf z}$.
        \While{not done}
            \State Sample a transitions $(s, a, r, s')$ from $\mathcal{R}$
            \State Obtain $\mu_s, \Sigma_s = \pi_{\psi}(s)$ 
            \State Calculate the log likelihood $\mathcal{J} = -\frac{1}{2} \log |\Sigma_s|-\frac{1}{2} \left(a-\mu_s\right)^{T} {\Sigma^{-1}_s}\left(a-\mu_s\right) + constant$
            \State $\psi \gets \psi + \nabla_{\psi}\mathcal{J}$
        \EndWhile
	\end{algorithmic}
	{\bf Output}: Initialized single-task policy $\pi_\psi$; inferred the task identity ${\bf z}$
\end{algorithm}

\begin{algorithm}[t]
    \caption{SAC initialized by our method}\label{alg:sac_our_method}
    \label{alg1}
    \textbf{Input}: unseen testing task $\mathcal{M}$; initialized single-task policy $\pi_\psi$; inferred the task identity ${\bf z}$; Q functions $Q_{\theta_1}, Q_{\theta_2}$ both initialized by $Q_D$ (from the multi-task policy); empty replay buffer $\mathcal{D}$
    \begin{algorithmic}[1]
        \State Set target parameters equal to main parameters $\theta_{\text{targ},1} \leftarrow \theta_1$, $\theta_{\text{targ},2} \leftarrow \theta_2$, $\psi_{\text{targ}} \leftarrow \psi$
        \Repeat
            \State Observe state $s$ and select action $a \sim \pi_{\psi}(\cdot|s)$
            \State Execute $a$ in task $\mathcal{M}$
            \State Observe next state $s'$, reward $r$, and done signal $d$ to indicate whether $s'$ is terminal
            \State Store $(s,a,r,s',d)$ in replay buffer $\mathcal{D}$
            \State If $s'$ is terminal, reset environment state.
            \If{it's time to update}
                \For{$j$ in range(however many updates)}
                    \For{$i=1,2$}
                        \State Randomly sample a batch of transitions, $B_i = \{ (s,a,r,s',d) \}$ from $\mathcal{D}$
                        \State Compute targets for the Q functions:
                        \begin{align*}
                            y (r,s',d) &= r + \gamma (1-d) \left(\min_{i=1,2} Q_{\theta_{\text{targ}, i}} (s', \tilde{a}', {\bf z}) - \alpha \log \pi_{\psi}(\tilde{a}'|s')\right), && \tilde{a}' \sim \pi_{\psi}(\cdot|s')
                        \end{align*}
                        \State Update Q-functions $i$ by one step of gradient descent using
                        \begin{align*}
                            & \nabla_{\theta_i} \frac{1}{|B_i|}\sum_{(s,a,r,s',d) \in B_i} \left( Q_{\theta_i}(s,a, {\bf z}) - y(r,s',d) \right)^2
                        \end{align*}
                        \EndFor
                    \State Using the transitions from $B_2$, update policy by one step of gradient ascent using
                    \begin{equation*}
                        \nabla_{\psi} \frac{1}{|B_2|}\sum_{s \in B_2} \Big(\min_{i=1,2} Q_{\theta_i}(s, \tilde{a}_{\psi}(s), {\bf z}) - \alpha \log \pi_{\psi} \left(\left. \tilde{a}_{\psi}(s) \right| s\right) \Big),
                    \end{equation*}
                    where $\tilde{a}_{\psi}(s)$ is a sample from $\pi_{\psi}(\cdot|s)$ which is differentiable wrt $\psi$ via the reparametrization trick \cite{kingma2013auto}.
                    \State Update target networks with
                    \begin{align*}
                        \theta_{\text{targ},i} &\leftarrow \rho \theta_{\text{targ}, i} + (1-\rho) \theta_i && \text{for } i=1,2\\
                        \psi_{\text{targ}} &\leftarrow \tau \psi_{\text{targ}} + (1-\tau) \psi
                    \end{align*}
                \EndFor
            \EndIf
        \Until{convergence}
    \end{algorithmic}
\end{algorithm}

Note that we perform imitation learning to initialize a new single-task policy instead of using the candidate action and perturbation generator to directly initialize the policy as the Q value function. To directly transfer the parameters of the candidate action and perturbation generator, we can initialize the action selection module with the distilled candidate action and perturbation generator and train them to generate action $a$ for state $s$ that maximizes the expected $Q(s, a)$ over a training batch. However, we find that this training procedures converge to a lower asymptotic performance.

\clearpage
\subsection{Comparison with a variation of SAC}\label{sec_init_counterpart}

In Sec. \ref{sec_exp_init_sac}, we show that SAC initialized by our method can significantly speed up the training on unseen testing tasks. An astute reader will notice that our implementation of Soft Actor Critic in Appendix \ref{sec_sac_our_methods} is different from the reference implementation. We maintain two identically initialized value functions by training them using different mini-batches. In the reference implementation, the two value functions are initialized differently but trained using the same mini-batch.

\begin{figure*}[t]
\centering
  \makebox[\textwidth]{
  
  \begin{subfigure}{0.20\paperwidth}
    \includegraphics[width=\linewidth]{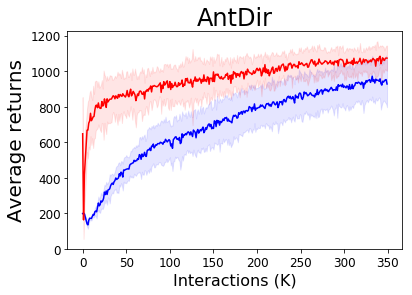}
  \end{subfigure}
  
  \begin{subfigure}{0.20\paperwidth}
    \includegraphics[width=\linewidth]{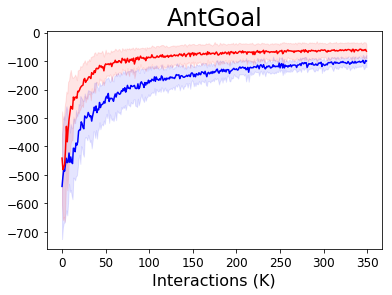}
  \end{subfigure}
    
  \begin{subfigure}{0.20\paperwidth}
    \includegraphics[width=\linewidth]{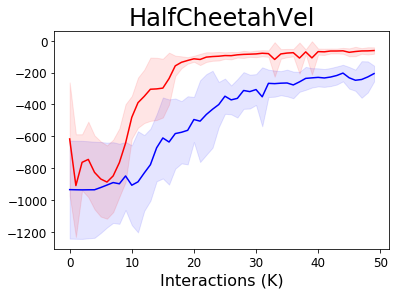}
  \end{subfigure}}
  
  \makebox[\textwidth]{
  \begin{subfigure}{0.20\paperwidth}
    \includegraphics[width=\linewidth]{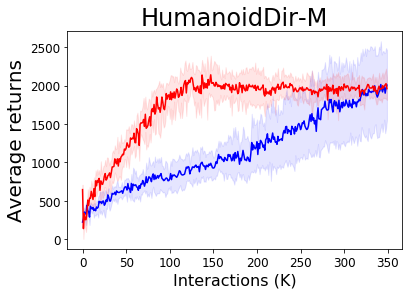}
  \end{subfigure}
  
  \begin{subfigure}{0.20\paperwidth}
    \includegraphics[width=\linewidth]{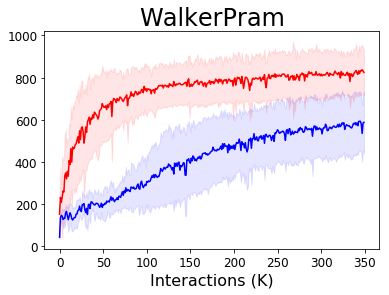}
  \end{subfigure}}
  \small{\color{red}--- }: SAC initialized by our multi-task policy \qquad  {\color{blue}--- }: A variation of SAC, randomly initialized 
  \caption{Comparison between the SAC initialized by our method and a variation of SAC. This variation trains two identically initialized Q functions with different mini-batches sampled from the replay buffer. Moreover, it also maintains a target policy network. The two methods share the same network sizes and architecture across all settings. In the figures above, the horizontal axis indicates the number of environment steps. The vertical axis indicates the average episode return. The shaded areas denote one standard deviation. }
\label{fig:sac_our_method_ablate}
\end{figure*}

To ensure the performance gain is really thanks to the good initialization provided by the multi-task policy, we compare its performance with a variation of SAC. Specifically, we initialize the two Q value functions identically but train them with different mini-batches sampled from the replay buffer. Moreover, we also maintain a target policy network as what is done in line 17 of Alg. \ref{alg:sac_our_method}. As shown in \autoref{fig:sac_our_method_ablate}, we can still observe that the SAC initialized by our methods outperform this variation of SAC. The unseen tasks used for evaluating the variation of SAC are the same as those used for testing the SAC initialized by our methods. The two methods share the same network sizes and architecture across all settings.

\clearpage

\section{Computing infrastructure and average run-time}

Our experiments are conducted on a machine with 2 GPUs and 8 CPUs. \autoref{table:runtime} provides the runtime for each of the experiment on all the task distributions. 

\begin{table}[H]
\vspace{0.2 in}
\centering
\small
\begin{tabular}{lccccccc}
\hline 
& AntDir & AntGoal & HDir & UGoal & HalfCheetahVel & WalkerParam\\
\hline 
Our full model  & 4.5 & 4.6 & 5.3 & 4.0 & 26.6 & 21.4\\ 

Contextual BCQ  & 5.9 & 5.9 & 5.3 & 5.5 & 17.2 & 13.1\\ 

PEARL & 8.2 & 7.6 & 8.1 & 6.9 & 16.5 & 11.1\\

No transition relabelling & 4.5 & 4.6 & 4.6 & 4.6 & 17.1 & 6.4\\ 

No triplet loss  & 4.4 & 5.6 & 4.1 & 4.6 & 17.4 & 8.2\\

Neither  & 4.5 & 4.6 & 3.9 & 6.2 & 16.7 & 7.5\\
SAC init by our method  & 4.2 & 4.3 & 5.3 & - &  0.6  & 4.2\\
\hline
\end{tabular}
\vspace{0.1 in}
\caption{Runtime of each experiment. The unit is hours. When calculating the runtime for algorithms that learn a multi-task policy, we exclude neither the time to generate the task buffers nor the time to learn single-task BCQ policies. The abbreviation HDir and UGoal stands for HumanoidDir-M and UmazeGoal-M, respectively. The runtime for SAC initialized by our methods is calculated by average across tasks from the corresponding task distribution.} 
\label{table:runtime}
\end{table}
\end{document}